\documentclass{article}
\usepackage{subfiles}

\usepackage{microtype}
\usepackage{graphicx}
\usepackage{subfigure}
\usepackage{booktabs}
\usepackage{amsmath}
\usepackage{amssymb}
\usepackage{amsfonts}
\usepackage{color,soul}
\usepackage{bm}
\usepackage{multirow}
\usepackage[english]{babel}

\newtheorem{corollary}{Corollary}

\usepackage{tikz}
\usetikzlibrary{arrows,shapes.arrows,chains}
\usepackage{verbatim}
\usetikzlibrary{bayesnet}
\tikzset{
    nand/.style = {-},
    and/.style = {->},
    ell/.style 2 args={
    ellipse,
    minimum width=1cm,
    minimum height=.5cm,
    draw,
    label={[name=#1]center:#2}
  },
  every loop/.style={}
}

\definecolor{bleudefrance}{rgb}{0.19, 0.55, 0.91}
\definecolor{brickred}{rgb}{0.8, 0.25, 0.33}
\definecolor{ceruleanblue}{rgb}{0.16, 0.32, 0.75}
\definecolor{darkolivegreen}{rgb}{0.33, 0.42, 0.18}

\usepackage{hyperref}

\usepackage[accepted]{icml2020}

\icmltitlerunning{Bayesian Graph Neural Networks with Adaptive Connection Sampling}

\begin{document}

\twocolumn[
\icmltitle{Bayesian Graph Neural Networks with Adaptive Connection Sampling}

\icmlsetsymbol{equal}{*}

\begin{icmlauthorlist}
\icmlauthor{Arman Hasanzadeh}{equal,to}
\icmlauthor{Ehsan Hajiramezanali}{equal,to}
\icmlauthor{Shahin Boluki}{to}
\icmlauthor{Mingyuan Zhou}{ed}
\icmlauthor{Nick Duffield}{to}
\icmlauthor{Krishna Narayanan}{to}
\icmlauthor{Xiaoning Qian}{to}
\end{icmlauthorlist}

\icmlaffiliation{to}{Electrical and Computer Engineering Department, Texas A\&M University, College Station, Texas, USA}
\icmlaffiliation{ed}{McCombs School of Business, The University of Texas at Austin, Austin, Texas, USA}

\icmlcorrespondingauthor{Arman Hasanzadeh}{armanihm@tamu.edu}

\icmlkeywords{Machine Learning, ICML}

\vskip 0.3in
]

\printAffiliationsAndNotice{\icmlEqualContribution} %

\begin{abstract}
We propose a unified framework for adaptive connection sampling in graph neural networks~(GNNs) that generalizes existing stochastic regularization methods for training GNNs. The proposed framework not only alleviates over-smoothing and over-fitting tendencies of deep GNNs, but also enables learning with uncertainty in graph analytic tasks with GNNs. %
Instead of using fixed sampling rates or hand-tuning them as model hyperparameters as in existing stochastic regularization methods, our adaptive connection sampling can be trained jointly with
GNN model parameters
in both global and local fashions. 
GNN training with adaptive connection sampling is shown to be mathematically equivalent to an efficient approximation of training Bayesian GNNs.  %
Experimental results with ablation studies on benchmark datasets validate that adaptively learning the sampling rate given graph training data is the key to %
boosting
the performance of GNNs in semi-supervised node classification, making them less prone to over-smoothing and over-fitting with more robust prediction.

\end{abstract}

\section{Introduction}

Graph neural networks (GNNs), and its numerous variants, have shown to be successful in graph representation learning by extracting high-level features for nodes from their topological neighborhoods. GNNs have boosted the state-of-the-art performance in a variety of graph analytic tasks, such as semi-supervised node classification and link prediction \citep{kipf2016semi, kipf2016variational,hasanzadeh2019sigvae, hajiramezanali2019vgrnn}.
Despite their successes, GNNs have two major limitations: 1) they cannot go very deep due to \textit{over-smoothing} and \textit{over-fitting} phenomena \citep{li2018deeper, kipf2016semi}; 2) the current implementations of GNNs do not provide uncertainty quantification~(UQ) of output predictions. 

There exist a variety of methods to address these problems.
For example, DropOut \citep{srivastava2014dropout} is %
a popular regularisation technique with deep neural networks~(DNNs) to avoid over-fitting, where network units are randomly masked during training. In GNNs, %
DropOut is realized by randomly removing the node features during training \citep{rong2019dropedge}. Often, the procedure is independent of the graph topology. However, empirical results have shown that, due to the nature of Laplacian smoothing in GNNs, graph convolutions have the over-smoothing tendency of mixing representations of adjacent nodes so that, when increasing the number of GNN layers, all nodes’ representations will converge to a stationary point, making them unrelated to node features
\citep{li2018deeper}. While it has been shown in~\citet{kipf2016semi} that DropOut alone is ineffectual in preventing over-fitting, partially due to over-smoothing, the combination of DropEdge, in which a set of edges are randomly removed from the graph, with DropOut has recently shown potential to alleviate these problems \citep{rong2019dropedge}.

On the other hand, with the development of efficient posterior computation algorithms, there have been successes in learning with uncertainty by Bayesian extensions of traditional deep network architectures, including convolutional neural networks~(CNNs). However, for GNNs, deriving their Bayesian extensions is more challenging due to their irregular neighborhood connection structures. 
In order to account for uncertainty in GNNs, \citet{zhang2019bayesian} present a Bayesian framework where the observed graph is viewed as a realization from a parametric family of random graphs. This allows joint inference of the graph and the GNN weights, leading to resilience to noise or adversarial attacks. Besides its prohibitive computational cost, the choice of the random graph model is important and can be inconsistent for different problems and datasets. Furthermore, the posterior inference in the current implementation only depends on the graph topology, but cannot consider node features.

In this paper, we introduce a general stochastic regularization technique for GNNs by adaptive connection sampling---Graph DropConnect (GDC). We show that existing GNN regularization techniques such as DropOut \citep{srivastava2014dropout}, DropEdge \citep{rong2019dropedge}, and node sampling \citep{chen2018fastgcn} are special cases of GDC. GDC regularizes neighborhood aggregation in GNNs at each channel, separately. This prevents connected nodes in graph from having the same learned representations in GNN layers; hence better improvement without serious over-smoothing can be achieved. Furthermore, adaptively learning the connection sampling or drop rate in GDC enables better stochastic regularization given graph data for target graph analytic tasks. In fact, our ablation studies show that only learning the DropEdge rate, without any DropOut, already substantially improves the performance in semi-supervised node classification with GNNs.  By probabilistic modeling of the \emph{connection} drop rate, we propose a hierarchical beta-Bernoulli construction for Bayesian learnable GDC, and derive the solution with both continuous relaxation and direct optimization with Augment-REINFORCE-Merge~(ARM) gradient estimates. With the naturally enabled UQ and regularization capability, our learnable GDC  can help address both over-smoothing and UQ challenges to further push the frontier of GNN research.

We further prove that adaptive connection sampling of GDC at each channel can be considered as random aggregation and diffusion in GNNs, with a similar Bayesian approximation interpretation as in Bayesian DropOut for CNNs \citep{gal2015bayesian}. 
Specifically, Monte Carlo estimation of GNN outputs can be used to evaluate the predictive posterior uncertainty. An important corollary of this formulation is that any GNN with neighborhood sampling, such as GraphSAGE \citep{hamilton2017inductive}, could be considered as its corresponding Bayesian approximation.%

\section{Preliminaries}
\subsection{Bayesian Neural Networks}
Bayesian neural networks (BNNs) %
aim to capture model uncertainty of DNNs by placing prior distributions over the model parameters %
to enable posterior updates during DNN training. It has been shown that these Bayesian extensions of traditional DNNs can be robust to over-fitting and provide appropriate prediction uncertainty estimation \citep{gal2016dropout,boluki2020learnable}.
Often, the standard Gaussian prior distribution is placed over the weights.
With random weights $\{\mathbf{W}^{(l)}\}_{l=1}^L$, the output prediction given an input $\mathbf{x}$ can be denoted by $\widehat{\mathbf{f}} \big( \mathbf{x}, \{\mathbf{W}^{(l)}\}_{l=1}^L \big)$, which is now a random variable in BNNs, enabling uncertainty quantification (UQ).

The key difficulty in using BNNs is that Bayesian inference is computationally intractable. There exist various methods that approximate BNN inference, such as Laplace approximation \citep{mackay1992bayesian}, sampling-based and stochastic variational inference \citep{paisley2012variational,rezende2014stochastic,hajiramezanali2020semi,dadaneh2020pairwise}, Markov chain Monte Carlo (MCMC) \citep{neal2012bayesian}, and stochastic gradient MCMC \citep{ma2015complete}.
However, their computational cost is still much higher than the non-Bayesian methods, due to the increased model complexity and slow convergence \citep{gal2016dropout}.%

\subsection{DropOut as Bayesian Approximation}
Dropout is commonly used in training many deep learning models as a way to avoid over-fitting. Using dropout at test time enables UQ with Bayesian interpretation of the network outputs as Monte Carlo samples of its predictive distribution \citep{gal2016dropout}. %
Various dropout methods have been proposed to multiply the output of each neuron by a random mask drawn from a desired distribution, such as Bernoulli \citep{hinton2012improving, srivastava2014dropout} and Gaussian \citep{kingma2015variational,srivastava2014dropout}. Bernoulli dropout and its extensions are the most commonly used in practice due to their ease of implementation and computational efficiency in existing deep architectures.

\subsection{Over-smoothing \& Over-fitting in GNNs}
It has been shown that graph convolution in graph convolutional neural networks (GCNs) \citep{kipf2016semi} is simply a special form of Laplacian smoothing, which mixes the features of a node and its nearby neighbors. Such diffusion operations lead to similar learned representations when the corresponding nodes are close topologically with similar features,
thus greatly improving node classification performance.
However, it also brings potential concerns of \textit{over-smoothing} \cite{li2018deeper}. If a GCN is deep with many convolutional layers, the learned representations may be over-smoothed and nodes with different topological and feature characteristics may become indistinguishable. More specifically, by repeatedly applying Laplacian smoothing many times, the node representations within each connected component of the graph will converge to the same values. %

Moreover, GCNs, like other deep models, may suffer from \textit{over-fitting} when we utilize an over-parameterized model to fit a distribution with limited training data, where the model we learn fits the training data very well but generalizes poorly to the testing data. 
\vspace{-.02 in}

\subsection{Stochastic Regularization \& Reduction for GNNs}
Quickly increasing model complexity and possible over-fitting and over-smoothing when modeling large graphs, as empirically observed in the 
GNN literature, have been conjectured for the main reason of limited %
performance
from deep GNNs \citep{kipf2016semi,rong2019dropedge}. Several stochastic regularization and reduction methods in GNNs have been proposed to improve the deep GNN performance. For example, \textit{stochastic regularization techniques}, such as DropOut \citep{srivastava2014dropout} and DropEdge \citep{rong2019dropedge}, have been used to prevent over-fitting and over-smoothing in GNNs. %
Sampling-based \textit{stochastic reduction by} random walk neighborhood sampling \citep{hamilton2017inductive} and node sampling \citep{chen2018fastgcn} %
has been deployed in GNNs to reduce the size of input data and thereafter model complexity. %
Next, we review each of these methods and show that they can be formulated in our proposed adaptive connection sampling  framework.

Denote the output of the $l$th hidden layer in GNNs by $\mathbf{H}^{(l)} = [\mathbf{h}^{(l)}_0, \dots, \mathbf{h}^{(l)}_n]^T\in \mathbb{R}^{n \times f_l}$ with $n$ being the number of nodes and $f_l$ being the number of output features at the $l$th layer. Assume $\mathbf{H}^{(0)} = \mathbf{X} \in \mathbb{R}^{n \times f_0}$ is the input matrix of node attributes, where $f_0$ is the number of nodes features. Also, assume that $\mathbf{W}^{(l)} \in \mathbb{R}^{f_l \times f_{l+1}}$ and $\sigma(\,\cdot\,)$ are the GNN parameters at the $l$th layer and the corresponding activation function, respectively. Moreover, $\mathcal{N}(v)$ denotes the neighborhood of node $v$; $\hat{\mathcal{N}}(v) = \mathcal{N}(v) \cup \{v\}$; and $\mathfrak{N}(.)$ is the normalizing operator, i.e., $\mathfrak{N}(\mathbf{A}) = \mathbf{I}_N + \mathbf{D}^{-1/2}\, \mathbf{A}\, \mathbf{D}^{-1/2}$. Finally, $\odot$ represents the Hadamard product.

\subsubsection{DropOut \citep{srivastava2014dropout}}
In a GNN layer, DropOut randomly removes output elements of its previous hidden layer $\mathbf{H}^{(l)}$ based on independent Bernoulli random draws with a constant success rate at each training iteration. This can be formulated as follows:
\begin{equation}
    \mathbf{H}^{(l+1)} = \sigma \left(\mathfrak{N}(\mathbf{A})(\mathbf{Z}^{(l)} \odot \mathbf{H}^{(l)})\, \mathbf{W}^{(l)}  \right),
\label{eq:dropout}
\end{equation}
where $\mathbf{Z}^{(l)}$ is a random binary matrix, with the same dimensions as $\mathbf{H}^{(l)}$, whose elements are samples of $\mathrm{Bernoulli}(\pi)$. Despite its success in fully connected and convolutional neural networks, DropOut has shown to be ineffectual in GNNs for preventing over-fitting and over-smoothing.

\subsubsection{DropEdge \citep{rong2019dropedge}}
DropEdge randomly removes edges from the graph by drawing independent Bernoulli random variables (with a constant rate) at each iteration. More specifically, a GNN layer with DropEdge can be written as follows:
\begin{equation}
    \mathbf{H}^{(l+1)} = \sigma \left(\mathfrak{N}(\mathbf{A} \odot \mathbf{Z}^{(l)})\, \mathbf{H}^{(l)} \, \mathbf{W}^{(l)}  \right),
\label{eq:dropedge}
\end{equation}
Note that here, the random binary mask, i.e. $\mathbf{Z}^{(l)}$, has the same dimensions as $\mathbf{A}$. Its elements are random samples of $\mathrm{Bernoulli}(\pi)$ where their corresponding elements in $\mathbf{A}$ are non-zero and zero everywhere else. It has been shown that the combination of DropOut and DropEdge reaches the best performance in terms of mitigating overfitting in GNNs.

\subsubsection{Node Sampling \citep{chen2018fastgcn}}
To reduce expensive computation in batch training of GNNs, due to the recursive expansion of neighborhoods across layers, \citet{chen2018fastgcn} propose to relax the requirement of simultaneous availability of test data. Considering graph convolutions as integral transforms of embedding functions under probability measures allows for the use of Monte Carlo approaches to consistently estimate the integrals. This leads to an optimal node sampling strategy, FastGCN, which can be formulated as
\vspace{-.03 in}
\begin{equation}
\label{eq:nodesample}
    \mathbf{H}^{(l+1)} = \sigma \left(\mathfrak{N}(\mathbf{A})\, \mathrm{diag}(\mathbf{z}^{(l)}) \mathbf{H}^{(l)} \, \mathbf{W}^{(l)}  \right),
\end{equation}
where $\mathbf{z}^{(l)}$ is a random vector whose elements are drawn from $\mathrm{Bernoulli}(\pi)$. This, indeed, is a special case of DropOut, as all of the output features for a node are either completely kept or dropped while DropOut randomly removes some of these related output elements associated with the node.

\section{Graph DropConnect}%
We propose a general stochastic regularization technique for GNNs---Graph DropConnect (GDC)---by adaptive connection sampling, which can be interpreted as an approximation of Bayesian GNNs.

In GDC, we allow GNNs to draw different random masks for each channel and edge independently. More specifically, the operation of a GNN layer with GDC is defined as follows:
\vspace{-.05 in}
\begin{gather}
    \mathbf{H}^{(l+1)}[:,j] = \sigma \left(\sum_{i=1}^{f_l}\mathfrak{N}(\mathbf{A} \odot \mathbf{Z}_{i,j}^{(l)})\, \mathbf{H}^{(l)}[:,i] \, \mathbf{W}^{(l)}[i,j]  \right),\nonumber\\
    \text{for}\quad j = 1,\, \dots \,, f_{l+1} 
    \label{equ: gdc}
\end{gather}
where $f_l$ and $f_{l+1}$ are the number of features at layers $l$ and $l+1$, respectively, and $\mathbf{Z}_{i,j}^{(l)}$ is a sparse random matrix (with the same sparsity as $\mathbf{A}$) whose non-zero elements are randomly drawn by $\mathrm{Bernoulli}(\pi_l)$. Note that $\pi_l$ can be different for each layer for GDC instead of assuming the same constant drop rate for all layers in previous methods.

As shown in~(\ref{eq:dropout}), (\ref{eq:dropedge}), and (\ref{eq:nodesample}), DropOut \cite{srivastava2014dropout}, DropEdge \citep{rong2019dropedge}, and Node Sampling \cite{chen2018fastgcn} have different sampling assumptions on channels, edges, or nodes, yet there is no clear evidence to favor one over the other in terms of consequent graph analytic performance. In the proposed GDC approach, there is a free parameter $\{\mathbf{Z}_{i,j}^{(l)} \in \{0, 1\}^{n \times n}\}_{i=1}^{f_l}$ to adjust the binary mask for the edges, nodes and channels. Thus the proposed GDC model has one extra degree of freedom to incorporate flexible connection sampling.

The previous stochastic regularization techniques can be considered as special cases of GDC. %
To illustrate that, we assume $\mathbf{Z}_{i,j}^{(l)}$ are the same for all $j \in \{1, 2, \dots, f_{l+1}\}$, thus we can omit the indices of the output elements at layer $l+1$ and rewrite~(\ref{equ: gdc}) as
\vspace{-.1 in}
\begin{gather}
    \mathbf{H}^{(l+1)} = \sigma \left(\sum_{i=1}^{f_l}\mathfrak{N}(\mathbf{A} \odot \mathbf{Z}_i^{(l)})\, \mathbf{H}^{(l)}[:,i] \, \mathbf{W}^{(l)}[i,:]  \right)
    \raisetag{22pt}
\label{equ: gdc_red}
\end{gather}
Define $\mathbf{J}_n$ as a $n \times n$ all-one matrix.
Let $\mathbf{Z}^{(l)}_{DO} \in \{0, 1\}^{n \times f_l}$, $\mathbf{Z}^{(l)}_{DE} \in \{0, 1\}^{n \times n}$, and $\mathrm{diag}(\mathbf{z}_{NS}^{(l)}) \in \{0, 1\}^{n \times n}$ be the random binary matrices corresponding to the ones adopted in DropOut \cite{srivastava2014dropout}, DropEdge \citep{rong2019dropedge}, and Node Sampling \cite{chen2018fastgcn}, respectively. The random mask $\{\mathbf{Z}_i^{(l)} \in \{0, 1\}^{n \times n}\}_{i=1}^{f_l}$ in GDC become the same as those of the DropOut when $\mathbf{Z}_i^{(l)} = \mathbf{J}_n \, \mathrm{diag}(\mathbf{Z}^{(l)}_{DO}[:,\,i])$, the same as those of DropEdge when $\{\mathbf{Z}_i^{(l)}\}_{i=1}^{f_l} = \mathbf{Z}^{(l)}_{DE}$, and  the same as those of node sampling when $\{\mathbf{Z}_i^{(l)}\}_{i=1}^{f_l} = \mathbf{J}_n \mathrm{diag}(\mathbf{z}_{NS}^{(l)})$.

\subsection{GDC as Bayesian Approximation}
In GDC, random masking is applied to the adjacency matrix of the graph to regularize the aggregation steps at each layer of GNNs. In %
existing Bayesian neural networks, the model parameters, i.e. $\mathbf{W}^{(l)}$, are considered random to enable Bayesian inference based on predictive posterior given training data \cite{gal2017concrete,boluki2020learnable}. %
Here, we show that connection sampling in GDC can be transformed from the output feature space to the parameter space so that it can be considered as appropriate Bayesian extensions of GNNs. 

First, we rewrite \eqref{equ: gdc_red} to have a node-wise view of a GNN layer with GDC. More specifically,
\vspace{-.07 in}
\begin{equation}
    \mathbf{h}^{(l+1)}_v = \sigma \left(\frac{1}{c_v}\big(\sum_{u \in \hat{\mathcal{N}}(v)} \mathbf{z}_{vu}^{(l)} \odot \mathbf{h}^{(l)}_u\big)\, \mathbf{W}^{(l)} \right),
\label{equ: gdc_nw}
\end{equation}
where $c_v$ is a constant derived from the degree of node $v$, and $\mathbf{z}_{vu}^{(l)} \in \{0, 1\}^{1 \times f_l}$ is the mask  row vector corresponding to connection between nodes $v$ and $u$ in three dimensional tensor $\mathcal{Z}^{(l)} = [\mathbf{Z}_1^{(l)},\dots,\mathbf{Z}_{f_l}^{(l)}]$. For brevity and without loss of generality, we ignore the constant $c_v$ in the rest of this section. We can rewrite and reorganize \eqref{equ: gdc_nw} to transform the randomness from sampling to the parameter space as
\vspace{-.1 in}
\begin{equation}
\begin{split}
    \mathbf{h}^{(l+1)}_v &=\sigma \left(\big(\sum_{u \in \hat{\mathcal{N}}(v)} \mathbf{h}^{(l)}_u\, \mathrm{diag}(\mathbf{z}_{vu}^{(l)})\big)\, \mathbf{W}^{(l)} \right)\\
    &=\sigma \left(\sum_{u \in \hat{\mathcal{N}}(v)} \mathbf{h}^{(l)}_u\, \big(\mathrm{diag}(\mathbf{z}_{vu}^{(l)})\, \mathbf{W}^{(l)}\big) \right). 
\end{split}
\label{equ: noise}
\end{equation}
Define $\mathbf{W}_{vu}^{(l)} := \mathbf{z}_{vu}^{(l)}\, \mathbf{W}^{(l)}$. We have:
\begin{equation}
    \mathbf{h}^{(l+1)}_v = \sigma \left(\sum_{u \in \hat{\mathcal{N}}(v)} \mathbf{h}^{(l)}_u\, \mathbf{W}_{vu}^{(l)} \right).
\label{equ: noise_last}
\end{equation}
$\mathbf{W}_{vu}^{(l)}$, which pairs the corresponding weight parameter with the edge in the given graph. The operation with GDC in \eqref{equ: noise_last} can be interpreted as learning different weights for each of the message passing along edges $e = (u, v) \in \mathcal{E}$ where $\mathcal{E}$ is the union of edge set of the input graph and self-loops for all nodes.

Following the variational interpretation in~\citet{gal2017concrete}, GDC can be seen as an approximating distribution $q_\theta(\boldsymbol{\omega})$ for the posterior $p(\boldsymbol{\omega} \, |\, \mathbf{A}, \mathbf{X})$ when considering a set of random weight matrices $\boldsymbol{\omega} = \{\boldsymbol{\omega}_e\}_{e=1}^{|\mathcal{E}|}$ in the Bayesian framework, where $\boldsymbol{\omega}_e = \{\mathbf{W}_{e}^{(l)}\}_{l=1}^L$ is the set of random weights for the $e$th edge, $|\mathcal{E}|$ is the number of edges in the input graph, and $\theta$ is the set of variational parameters. 
The Kullback--Leibler (KL) divergence $\mathrm{KL}(q_\theta(\boldsymbol{\omega}) || p(\boldsymbol{\omega}))$ is considered in training as a regularisation term, which ensures that the approximating  $q_\theta(\boldsymbol{\omega})$ does not deviate too far from the prior distribution. To be able to evaluate the KL term analytically, the discrete quantised Gaussian can be adopted as the prior distribution as in \citet{gal2017concrete}. Further with the factorization  $q_\theta(\boldsymbol{\omega})$ over $L$ layers and $|\mathcal{E}|$ edges  such that $q_\theta(\boldsymbol{\omega}) = \prod_l \prod_e q_{\theta_l}(\mathbf{W}_{e}^{(l)})$ and letting $q_{\theta_l}(\mathbf{W}_{e}^{(l)}) = \pi_l \delta(\mathbf{W}_{e}^{(l)}- 0) + (1 - \pi_l) \delta(\mathbf{W}_{e}^{(l)}- \mathbf{M}^{(l)})$, where $\theta_l = \{\mathbf{M}^{(l)}, \pi_l\}$, the KL term can be written as $\sum_{l=1}^L \sum_{e=1}^{|\mathcal{E}|} \mathrm{KL}(q_{\theta_l}(\mathbf{W}_{e}^{(l)}) \, ||\, p(\mathbf{W}_{e}^{(l)}))$ and approximately
\begin{equation*}
    \mathrm{KL}(q_{\theta_l}(\mathbf{W}_{e}^{(l)}) \,||\, p(\mathbf{W}_{e}^{(l)})) \propto \frac{(1-\pi_l)}{2} \,||\, \mathbf{M}^{(l)} ||^2 - \mathcal{H}(\pi_l),
\end{equation*}
where $\mathcal{H}(\pi_l)$ is the entropy of a Bernoulli random variable with the success rate $\pi_l$. 

Since the entropy term does not depend on network weight parameters $\mathbf{M}^{(l)}$, it can be omitted when $\pi_l$ is not optimized. But we learn $\pi_l$ in GDC, thus the entropy term is important.  %
Minimizing the KL divergence with respect to the drop rate $\pi_l$ is equivalent to maximizing the entropy of a Bernoulli random variable with probability $1 - \pi_l$. This pushes the drop rate towards 0.5, which may not be desired in some cases where higher/lower drop rate probabilities are more appreciated. %

\subsection{Variational Inference for GDC}

We consider $\mathbf{z}_e^{(l)}$ and $\mathbf{W}^{(l)}$ as local and global random variables, respectively, and denote $\mathbf{Z}^{(l)} = \{\mathbf{z}_e^{(l)}\}_{e=1}^{|\mathcal{E}|}$ and $\boldsymbol{\omega}^{(l)} = \{\mathbf{W}^{(l)}_e\}_{e=1}^{|\mathcal{E}|}$. For  inference of this approximating model with GDC, we assume a factorized variational distribution $q(\boldsymbol{\omega}^{(l)}, \mathbf{Z}^{(l)}) = q(\boldsymbol{\omega}^{(l)}) \, q(\mathbf{Z}^{(l)})$.  
Let the prior distribution  $p(\mathbf{W}_e^{(l)})$ be a discrete quantised Gaussian and $p(\boldsymbol{\omega}^{(l)}) = \prod_{e=1}^{\mathcal{E}} p(\mathbf{W}_e^{(l)})$.%
Therefore, the KL term can be written as $\sum_{l=1}^L \mathrm{KL}(q(\boldsymbol{\omega}^{(l)}, \mathbf{Z}^{(l)}) \, || \, p(\boldsymbol{\omega}^{(l)}, \mathbf{Z}^{(l)}))$, %
with
\begin{equation}
\begin{split}
&\mathrm{KL}\left(q(\boldsymbol{\omega}^{(l)} , \mathbf{Z}^{(l)}) \, || \, p(\boldsymbol{\omega}^{(l)}, \mathbf{Z}^{(l)}) \right) \propto \nonumber \\
& \quad \frac{|\mathcal{E}| (1-\pi_l) }{2} ||\mathbf{M}^{(l)}||^2 + \, \sum_{e=1}^{|\mathcal{E}|} \mathrm{KL}\left(q(\mathbf{z}_{e}^{(l)}) \, || \,  p(\mathbf{z}_{e}^{(l)})\right). \nonumber
\end{split}
\end{equation}
The KL term consists of the common weight decay in the non-Bayesian GNNs with the additional KL term $\sum_{e=1}^{|\mathcal{E}|} \, \mathrm{KL}(q(\mathbf{z}_{e}^{(l)}) \, || \, p(\mathbf{z}_{e}^{(l)}))$ that acts as a regularization term for $\mathbf{z}_{e}^{(l)}$. %
In this GDC framework, the variational inference loss, for node classification for example, can be written~as
\begin{equation}
\begin{split}
&\mathcal{L}(\{\mathbf{M}^{(l)},\pi_l\}_{l=1}^{L}) =\\
&\quad\mathbb{E}_{q(\{\boldsymbol{\omega}^{(l)} , \mathbf{Z}^{(l)}\}_{l=1}^{L} )}[\mathrm{log}P(Y_{o}|X,\{\boldsymbol{\omega}^{(l)}, \mathbf{Z}^{(l)}\}_{l=1}^{L})] \\
&\quad - \sum_{l=1}^L \mathrm{KL}(q(\boldsymbol{\omega}^{(l)}, \mathbf{Z}^{(l)}) \, || \, p(\boldsymbol{\omega}^{(l)}, \mathbf{Z}^{(l)})),
\end{split}
\raisetag{21pt}
\label{eq:varloss}
\end{equation}
where $Y_{o}$ denotes the collection of the available labels for the observed nodes.
The optimization of \eqref{eq:varloss} with respect to the weight matrices can be done by a Monte Carlo sample, i.e. sampling a random GDC mask and calculating the gradients with respect to $\{\mathbf{M}^{(l)}\}_{l=1}^{L}$ with stochastic gradient descent. It is easy to see that if $\{\pi_l\}_{l=1}^{L}$ are fixed, implementing our GDC is as simple as using common regularization terms on the neural network weights.

We aim to optimize the drop rates $\{\pi_l\}_{l=1}^{L}$ jointly with the weight matrices. This clearly provides more flexibility as all the parameters of the approximating posterior will be learned from the data instead of being fixed \emph{a priori} or treated as hyper-parameters, often difficult to tune. However, the optimization of \eqref{eq:varloss} with respect to the drop rates is challenging. Although the KL term is not a function of the random masks, the commonly adopted reparameterization techniques \citep{rezende2014stochastic,kingma2013auto} are not directly applicable here for computing the expectation in the first term since the drop masks are binary. Moreover, score-function
gradient estimators, such as REINFORCE \cite{williams1992simple, fu2006gradient}, possess high variance.
One potential solution is continuous relaxation of the drop masks. This approach has lower variance at the expense of introducing bias. Another solution is the direct optimization with respect to the discrete variables by the recently developed Augment-REINFORCE-Merge~(ARM) method \cite{yin2018arm}, which has been used in BNNs \cite{boluki2020learnable} and information retrieval \cite{icassp_arsm,dadaneh2020pairwise}. In the next section, we will discuss in detail about our GDC formulation with more flexible beta-Bernoulli prior construction for adaptive connection sampling and how we solve the joint optimization problem for training GNNs with adaptive connection sampling.
 
\section{Variational Beta-Bernoulli GDC}

The sampling or drop rate in GDC can be set as a constant hyperparameter as commonly done in other stochastic regularization techniques. In this work, we further enrich GDC with an adaptive sampling mechanism, where the drop rate is directly learned together with GNN parameters given graph data. In fact, in the Bayesian framework, such a hierarchical construct may increase the model expressiveness to further improve prediction and uncertainty estimation performance, as we will %
show empirically in Section~\ref{sec:exp}.

Note that in this section, for brevity and simplicity we do the derivations for one feature dimension only, i.e. $f_l=1$. Extending to multi-dimensional features is straightforward as we assume the drop masks are independent across features. Therefore, we drop the feature index in our notations.
Inspired by the beta-Bernoulli process \citep{thibaux2007hierarchical}, whose marginal representation is also known as the Indian Buffet Process~(IBP) \citep{ghahramani2006infinite}, we impose a beta-Bernoulli prior to the binary random masks as%
\begin{eqnarray}
a_{e}^{(l)} &=& z_e^{(l)} a_e , \quad 
    z_{e}^{(l)} \sim  \mathrm{Bernoulli}(\pi_l), \nonumber \\
    \pi_l &\sim& \mathrm{Beta}(c/L, \, c (L - 1)/L),
\label{eq:beta-bernoulli}
\end{eqnarray}
where $a_e$ denotes an element of the adjacency matrix $\mathbf{A}$ corresponding to an edge $e$, and $\hat{a}_e^{(l)}$ an element of the matrix $\hat{\mathbf{A}}^{(l)} = \mathbf{A} \odot \mathbf{Z}^{(l)}$. Such a formulation is known to be capable of enforcing sparsity in random masks \citep{zhou2009non,hajiramezanali2018bayesian}, which has been shown to be necessary for regularizing deep GNNs as discussed in DropEdge~\cite{rong2019dropedge}.

With this hierarchical beta-Bernoulli GDC formulation, inference based on Gibbs sampling can be computationally demanding for large datasets, including graph data \citep{hasanzadeh2019sigvae}. In the following, we derive efficient variational inference algorithm(s) for learnable GDC.

To perform variational inference for GDC random masks and the corresponding drop rate at each GNN layer together with weight parameters, we define the variational distribution as $q(\mathbf{Z}^{(l)}, \pi_l) = q(\mathbf{Z}^{(l)} \,|\, \pi_l) q(\pi_l)$.
We define $q(\pi_l)$ to be Kumaraswamy distribution \cite{kumaraswamy1980generalized}; as an alternative to the beta prior factorized over $l$th layer
\begin{equation}
    q(\pi_l; a_l, b_l) = a_l b_l \pi_l^{a_l -1}(1 - \pi_l^{a_l})^{b_l - 1},
\end{equation}
where $a_l$ and $b_l$ are greater than zero. Knowing $\pi_l$ the edges are independent, thus we can rewrite $q(\mathbf{Z}^{(l)} \,|\, \pi_l) = \prod_{e=1}^{|\mathcal{E}|}q(z_e^{(l)} \,|\, \pi_l)$. We further put a Bernoulli distribution with parameter $\pi_l$ over $q(\mathbf{z}^{(l)}_{e}|\pi_l)$. The KL divergence term can be written as
\begin{equation}
\begin{split}
&\mathrm{KL}\left(q(\mathbf{Z}^{(l)}, \pi_l) \,||\, p(\mathbf{Z}^{(l)}, \pi_l)\right) = \nonumber \\ 
&\,\,\,\sum_{e=1}^{|\mathcal{E}|} \mathrm{KL}\left(q(z_e^{(l)} \,|\, \pi_l) \,||\, p(z_e^{(l)} \,|\, \pi_l)\right) + \mathrm{KL}\left(q(\pi_l) \,||\, p(\pi_l)\right). \nonumber
\end{split}
\end{equation}
While the first term is zero due to the identical distributions, the second term can be computed in closed-form as
\begin{equation}
\begin{split}
&\mathrm{KL}\left(q(\pi_l) \,||\, p(\pi_l)\right) = \nonumber \\
&\quad\frac{a_l - c/L}{a_l}\left(- \gamma - \Psi(b_l) - \frac{1}{b_l} \right) + \mathrm{log} \frac{a_l b_l}{c/L} - \frac{b_l - 1}{b_l},
\end{split}
\end{equation}
where $\gamma$ is the Euler-Mascheroni constant and $\Psi(\cdot)$ is the digamma function. 

The gradient of the KL term in \eqref{eq:varloss} can easily be calculated with respect to the drop parameters. However, as mentioned in the previous section, due to the discrete nature of the random masks, we cannot directly apply reparameterization technique to calculate the gradient of the first term in \eqref{eq:varloss} with respect to the drop rates (parameters). One way to address this issue is to replace the discrete variables with a continuous approximation. We impose a concrete distribution relaxation \citep{jang2016categorical,gal2017concrete} for the Bernoulli random variable $z^{(l)}_{uv}$, leading to an efficient optimization by 
sampling from simple sigmoid distribution which has a
convenient parametrization
\begin{equation}
    \tilde{z}^{(l)}_{e} = \mathrm{sigmoid}\left(\frac{1}{t} \mathrm{log}\big(\frac{\pi_l}{1 - \pi_l}\big) + \mathrm{log}\big(\frac{u}{1-u}\big)\right),
\end{equation}
where $u \sim \mathrm{Unif}[0, 1]$ and $t$ is temperature parameter of relaxation. We can then use stochastic gradient variational Bayes to optimize the variational parameters $a_l$ and $b_l$.

Although this approach is simple, the relaxation introduces bias. Our other approach is to directly optimize the variational parameters using the original Bernoulli distribution in the formulation as in \citet{boluki2020learnable}. We can calculate the gradient of the variational loss with respect to $\boldsymbol{\alpha}=\{\text{logit}(1-\pi_l)\}_{l=1}^{L}$ using ARM estimator , which is unbiased and has low variance, by performing two forward passes as %
\vspace{-0.05cm}
\begin{equation}
\begin{split}
    \nabla_{\boldsymbol{u}}\mathcal{L}&(\mathbf{\alpha}) = \mathbb{E}_{u \sim \prod_{l=1}^{L}\prod_{e=1}^{|\mathcal{E}|}\mathrm{Unif}[0, 1](u_e^{(l)})}\Big[\big( \mathcal{L}(\{\mathbf{M}^{(l)}\}_{l=1}^L,\\ &1_{[\boldsymbol{u}>\sigma(-\boldsymbol{\alpha})]})-\mathcal{L}(\{\mathbf{M}^{(l)}\}_{l=1}^L,1_{[\boldsymbol{u}<\sigma(\boldsymbol{\alpha})]})\big) \big(\boldsymbol{u} - \frac{1}{2} \big) \Big],\nonumber
    \end{split}
\end{equation}
where $\mathcal{L}(\{\mathbf{M}^{(l)}\}_{l=1}^L,1_{[\boldsymbol{u}<\sigma(\boldsymbol{\alpha})]})$ denotes the loss obtained by setting 
$\mathbf{Z}^{(l)}=1_{[\boldsymbol{u}^{(l)}<\sigma(\boldsymbol{\alpha}_l)]}:=\big(1_{[u^{(l)}_1<\sigma(\alpha_l)]},\ldots, 1_{[u^{(l)}_{|\mathcal{E}|}<\sigma(\alpha_l)]}\big)$ for $l=1,\ldots, L$. The gradient with respect to $\{a_l,b_l\}_{l=1}^{L}$ can then be calculated by using the chain rule and the reparameterization for $\pi_l=(1-u^{\frac{1}{b_l}})^{\frac{1}{a_l}}, u\sim \text{Unif}[0,1]$. 

It is worth noting that although the beta-Bernoulli DropConnect with ARM is expected to provide better performance due to the more accurate gradient estimates, it has slightly higher computational complexity as it requires two forward passes.

\begin{table*}[!t]
    \caption{Semi-supervised node classification accuracy of GCNs with our adaptive connection sampling and baseline methods.}
    \label{tab:results}
    \centering
    \resizebox{2.0\columnwidth}{!}{
    \begin{tabular}{@{}l | c c c c c c}
    \toprule
        \multirow{1}{*}{\textbf{Method}}   & \multicolumn{2}{c}{\textbf{Cora}}     & \multicolumn{2}{c}{\textbf{Citeseer}}   & 
        \multicolumn{2}{c}{\textbf{Cora-ML}} \\
        & \multicolumn{1}{c}{2 layers} &  \multicolumn{1}{c}{4 layers} & \multicolumn{1}{c}{2 layers} & \multicolumn{1}{c}{4 layers} & \multicolumn{1}{c}{2 layers} & \multicolumn{1}{c}{4 layers} \\
        \midrule
        GCN-DO & $80.98\pm0.48$ & $78.24\pm2.4$  & $70.44\pm0.39$ & $64.38\pm0.90$ & $83.45\pm0.73$ & $81.51\pm1.01$ \\
        GCN-DE & $78.36\pm0.92$ & $73.40\pm2.07$ & $70.52\pm0.75$ & $57.14\pm0.90$ & $83.30\pm1.37$ & $68.89\pm3.37$ \\
        GCN-DO-DE & $80.58\pm1.19$ & $79.20\pm1.07$ & $70.74\pm1.23$ & $64.84\pm0.98$ & $83.61\pm0.83$ & $81.21\pm1.53$ \\
        \midrule
        \textbf{GCN-BBDE} & $\mathbf{81.58}\pm0.49$ & $\mathbf{80.42}\pm0.25$ & $\mathbf{71.46}\pm0.55$ & $\mathbf{68.58}\pm0.88$ & $\mathbf{84.62}\pm1.70$ & $\mathbf{84.73}\pm0.52$ \\
        \textbf{GCN-BBGDC} & $\mathbf{81.80}\pm0.99$ & $\mathbf{82.20}\pm0.92$ & $\mathbf{71.72}\pm0.48$ & $\mathbf{70.00}\pm0.36$ & $\mathbf{85.43}\pm0.70$ & $\mathbf{85.52}\pm0.83$ \\
        \bottomrule
    \end{tabular}
  }
\end{table*}

\section{Connection to Random Walk Sampling} 
Various types of random walk have been used in graph representation learning literature to reduce the size of input graphs. In GNNs, specifically in GraphSAGE \citep{hamilton2017inductive}, random walk sampling has been deployed to reduce the model complexity for very large graphs. One can formulate a GNN layer with random walk sampling as follows:
\vspace{-.02in}
\begin{equation}
    \mathbf{h}^{(l+1)}_v = \sigma \left(( \sum_{u \in \hat{\mathcal{N}}(v)} (z_{vu}^{(l)}\, |\, \mathbf{Z}^{(l-1)})\, \mathbf{h}^{(l)}_u )\, \mathbf{W}^{(l)} \right).
\end{equation}
Here, $\mathbf{Z}^{(l)}$ is the same as the one in DropEdge except that it is dependent on the masks from the previous layer. This is due to the fact that random walk samples for each node are connected subgraphs. %

In this setup, we can decompose the variational distribution of the GDC formulation in an autoregressive way. Specifically, here we have $q(z_{uv}^{(l)} | \mathbf{Z}^{(l-1)}) = \mathrm{Bernoulli}(\pi_l) 1_{\sum_{u \in \hat{\mathcal{N}}(v)} z_{vu}^{(l-1)}>0}$. With fixed Bernoulli parameters, we can calculate the gradients for the weight matrices with Monte Carlo integration. Learning Bernoulli parameters is challenging and does not allow direct application of ARM due to the autoregressive structure of the variational posterior. We leave sequential ARM for future study.
\vspace{-.1 in}

\begin{corollary}
Any graph neural network with random walk sampling, such as GraphSAGE, is an approximation of a Bayesian graph neural network as long as outputs are calculated using Monte Carlo sampling.
\end{corollary}

\section{Sampling Complexity}\label{sec:samp_comp}
The number of random samples needed for variational inference in GDC, \eqref{equ: gdc}, at each layer of a GNN is $|\mathcal{E}|\times f_{l} \times f_{l+1}$. This number would reduce to $|\mathcal{E}|\times f_{l}$ in the constrained version of GDC as shown in \eqref{equ: gdc_red}. These numbers, potentially, could be very high specially if the size of the graph or the number of filters are large, which could increase the space complexity and computation time. To circumvent this issue, we propose to draw a single sample for a block of features as oppose to drawing a new sample for every single feature. This would reduce the number of required samples to $|\mathcal{E}|\times \mathrm{nb}$ with $\mathrm{nb}$ being the number of blocks. In our experiments, we have one block in the first layer and two blocks in layers after that. 
In our experiments, we keep the order of features the same as the original input files, and divide them into $\mathrm{nb}$ groups with the equal number of features.

While in our GDC formulation, as shown in \eqref{equ: gdc} and \eqref{equ: gdc_red}, the normalization $\mathfrak{N}(\cdot)$ is applied after masking, one can multiply the randomly drawn mask with the pre-computed normalized adjacency matrix. This relaxation reduces the computation time and has negligible effect on the performance based on our experiments. An extension to the GDC sampling strategy is asymmetric sampling where the mask matrix $\mathbf{Z}$ could be asymmetric. This would increase the number of samples by a factor of two; however it increases the model flexibility. In our experiments, we have used asymmetric masks and multiplied the mask with the normalized adjacency matrix.

\section{Numerical Results}
\label{sec:exp}

We test the performance of our 
adaptive connection sampling framework, learnable GDC, on semi-supervised node classification using real-world citation graphs. In addition, we compare the uncertainty estimates of predictions by Monte Carlo beta-Bernoulli GDC and Monte Carlo Dropout. We also show the performance of GDC compared to existing methods in alleviating the issue of over-smoothing in GNNs. Furthermore, we investigate the effect of the number of blocks on the performance of GDC. We have also investigated learning separate drop rates for every edge in the network, i.e. \emph{local} GDC, which is included in the supplementary materials.

\subsection{Semi-supervised Node Classification}
\subsubsection{Datasets and Implementation Details}
We conducted extensive experiments for semi-supervised node classification with  real-world citation datasets.
We consider \emph{Cora}, \emph{Citeseer} 
and \emph{Cora-ML} 
datasets, and preprocess and split them same as \citet{kipf2016semi} and \citet{bojchevski2018deep}. We train beta-Bernoulli GDC (BBGDC) models for 2000 epochs with a learning rate of 0.005 and a validation set used for early stopping. All of the hidden layers in our implemented GCNs have 128 dimensional output features. We  use $5\times10^{-3}$, $10^{-2}$, and $10^{-3}$ as L2 regularization factor for Cora, Citeseer, and Cora-ML, respectively. For the GCNs with more than 2 layers, we 
use warm-up during the first 50 training epochs to gradually impose the beta-Bernoulli KL term in the objective function.
The temperature in the concrete distribution is set to 0.67. For a fair comparison, the number of hidden units are the same in the baselines and their hyper-parameters are hand-tuned to achieve their best performance. Performance is reported by the average accuracy with standard deviation based on 5 runs on the test set. The dataset statistics as well as more implementation details are included in the supplementary materials.

\subsubsection{Discussion}
Table \ref{tab:results} shows that BBGDC outperforms the state-of-the-art stochastic regularization techniques in terms of accuracy in all benchmark datasets. DO and DE in the table stand for DropOut and DropEdge, respectively.
Comparing GCN-DO and GCN-DE, we can see that DropEdge alone is less effective than DropOut in overcoming over-smoothing and over-fitting in GCNs. The difference between accuracy of GCN-DO and GCN-DE is more substantial in deeper networks (5\% in Cora, 7\% in Citeseer, and 13\% in Cora-ML),  which further proves the limitations of DE. Among the baselines, combination of DO and DE shows the best performance allowing to have deeper models. However, this is not always true. For example in Citeseer, 4-layer GCN shows significant decrease in performance compared to 2-layer GCN.

To show the advantages of learning the drop rates as well as the effect of hierarchical beta-Bernoulli construction, we have also evaluated beta-Bernoulli DropEdge (BBDE) with the concrete approximation, in which the edge drop rate at \emph{each layer} is learned using the same beta-Bernoulli hierarchical construction as GDC. We see that GCN with BBDE, without any DropOut, performs better than both GCNs with DE and DO-DE. By comparing GCN with BBDE and GCN with BBGDC, it is clear that the improvement is not only due to learnable sampling rate but also the increased flexibility of GDC compared to DropEdge. We note that GCN-BBGDC is the only method for which the accuracy improves by increasing the number of layers except in Citeseer.

\subsubsection{Concrete relaxation versus ARM}
To investigate the effect of direct optimization of the variational loss with respect to the drop parameters with ARM vs relaxation of the discrete random variables with concrete, we construct three ARM optimization-based variants of our method: Learnable Bernoulli DropEdge with ARM gradient estimator (BDE-ARM) where the edge drop rate of the Bernoulli mask at each layer is directly optimized; beta-Bernoulli DropEdge with ARM (BBDE-ARM); and beta-Bernoulli GDC with ARM (BBGDC-ARM). We evaluate these methods on the 4-layer GCN setups where significant performance improvement compared with the baselines has been achieved by BBDE and GDC with concrete relaxation. Comparing the performance of BBDE-ARM and BBGDC-ARM in Table \ref{tab: acc_ssnc} with the corresponding models with concrete relaxation, suggests further improvement when the drop parameters are directly optimized. Moreover, BDE-ARM, which optimizes the parameters of the Bernoulli drop rates, performs better than DO, DE, and DO-DE.

\begin{table}[!t]
   \caption{Accuracy of ARM optimization-based variants of our proposed method in semi-supervised node classification.}
  
\vspace{0.25em}
  \centering
   \resizebox{1.0\columnwidth}{!}{
  \begin{tabular}{@{}l | c c c }
    \toprule
     {\textbf{Method}} & {\textbf{Cora} (4 layers)} & {\textbf{Citeseer} (4 layers)} \\
    \midrule
    \textbf{GCN-BDE-ARM} & $79.95\pm0.79$ & $67.90\pm0.15$\\
    \textbf{GCN-BBDE-ARM} & $81.78\pm0.81$ & $69.43\pm0.45$\\
    \textbf{GCN-BBGDC-ARM} & $82.40\pm0.60$ & $70.25\pm0.07$\\
    \bottomrule
  \end{tabular}
   }
  \label{tab: acc_ssnc}
  \vspace{-0.2in}
\end{table}

\subsection{Uncertainty Quantification}
To evaluate the quality of uncertainty estimates obtained by our model, we use the Patch Accuracy vs Patch Uncertainty (PAvPU) metric introduced in \cite{mukhoti2018evaluating}. PAvPU combines $p(\mathrm{accurate|certain})$, i.e. the probability that the model is accurate on its output given that it is confident on the same, $p(\mathrm{certain|inaccurate})$, i.e. the probability that the model is uncertain about its output given that it has made a mistake in its prediction, into a single metric. More specifically, it is defined as $\mathrm{PAvPU} = (n_{ac}+n_{iu})/(n_{ac}+n_{au}+n_{ic}+n_{iu}),$
where $n_{ac}$ is the number of accurate and certain predictions, $n_{au}$ is the number of accurate and uncertain predictions, $n_{ic}$ is the number of inaccurate and certain predictions, and $n_{iu}$ is the number of inaccurate and uncertain predictions. Higher PAvPU means that certain predictions are accurate and inaccurate predictions are uncertain.

\begin{figure}[!t]
    \centering
    \includegraphics[width=1.0\columnwidth,keepaspectratio,trim={0 0 1cm 1cm},clip]{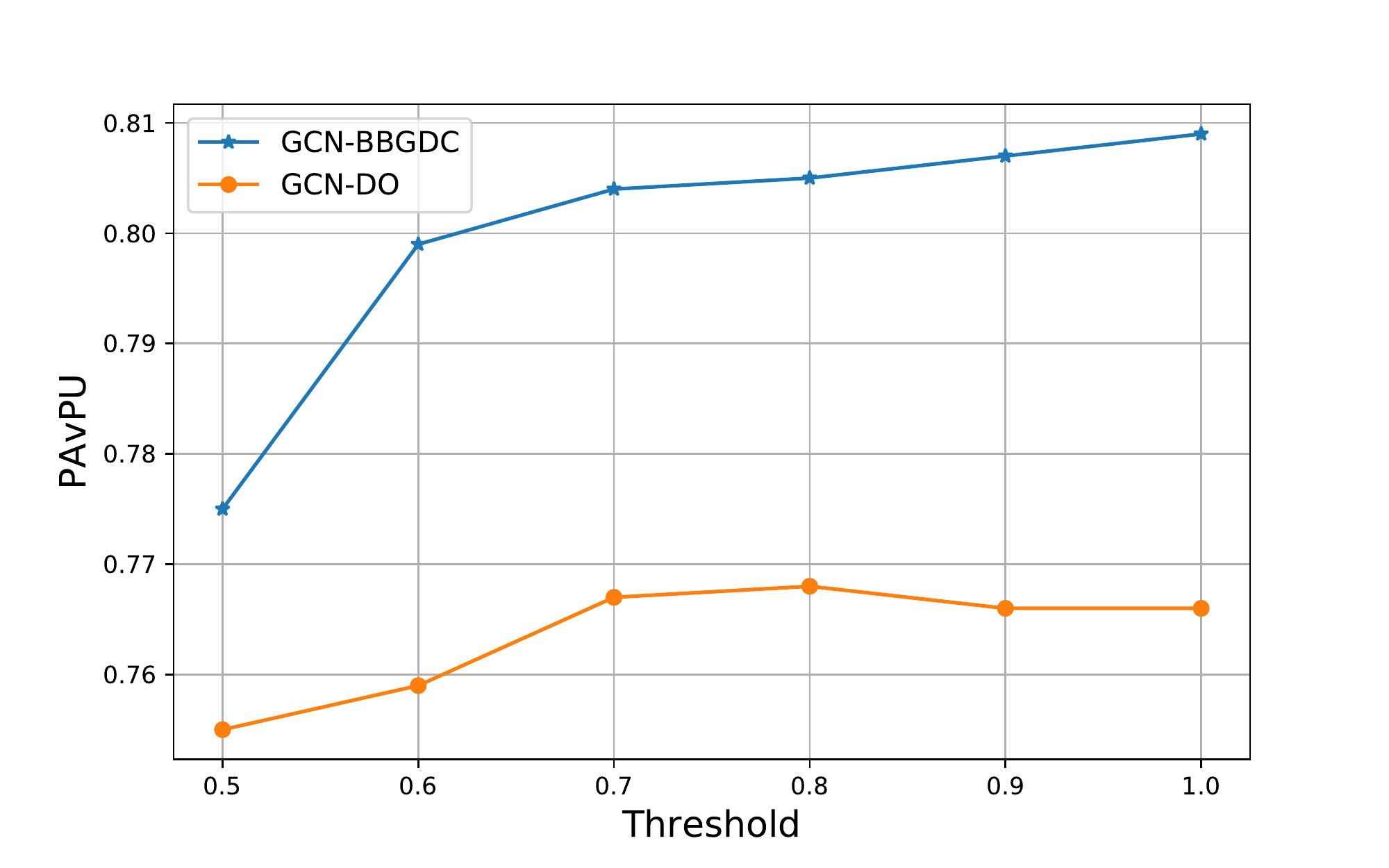}
    \vspace{-0.2in}
    \caption{Comparison of uncertainty estimates in PAvPU by a 4-layer GCN-BBGDC with 128-dimensional hidden layers and a 4-layer GCN-DO 128-dimensional hidden layers on Cora.}%
    \label{fig: uncertain}
\end{figure}

We here demonstrate the results for uncertainty estimates for a 4-layer GCN-DO and a 4-layer GCN-BBGDC with random initialization for semi-supervised node classification on Cora. We have evaluated PAvPU using 20 Monte Carlo samples for the test set where we use predictive entropy as the uncertainty metric. The results are shown in Figure \ref{fig: uncertain}. It can be seen that our proposed model consistently outperforms GCN-DO on every uncertainty threshold ranging from 0.5 to 1 of the maximum predictive uncertainty. While Figure \ref{fig: uncertain} depicts the results based on one random initialization, other initializations show the same trend.

\begin{figure*}[!t]
    \centering
    \begin{subfigure}{}
        \includegraphics[width=.47\textwidth,trim={0 0 1cm 1cm},clip]{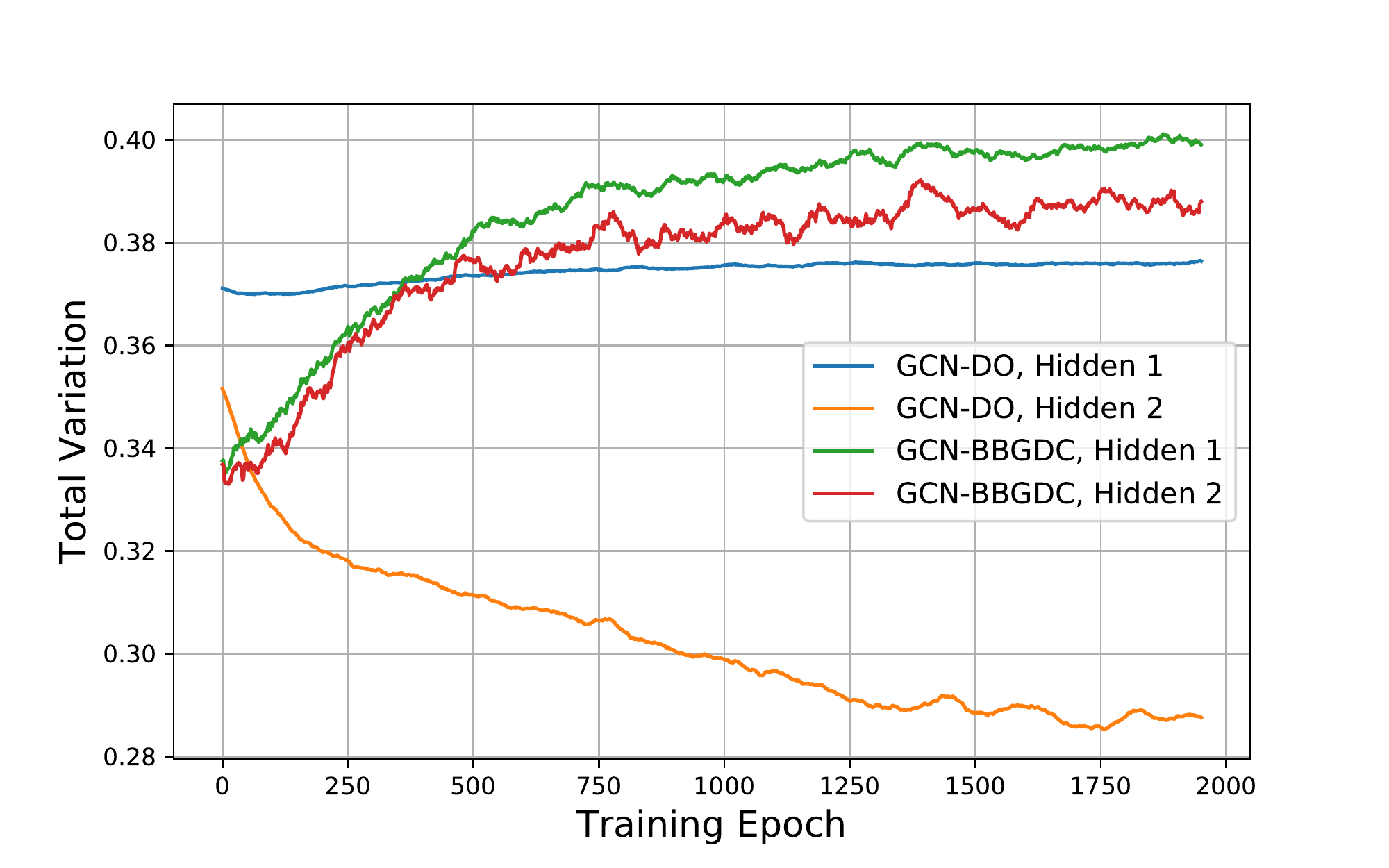}
    \end{subfigure}
    \begin{subfigure}{}
        \includegraphics[width=.47\textwidth,trim={0 0 1cm 1cm},clip]{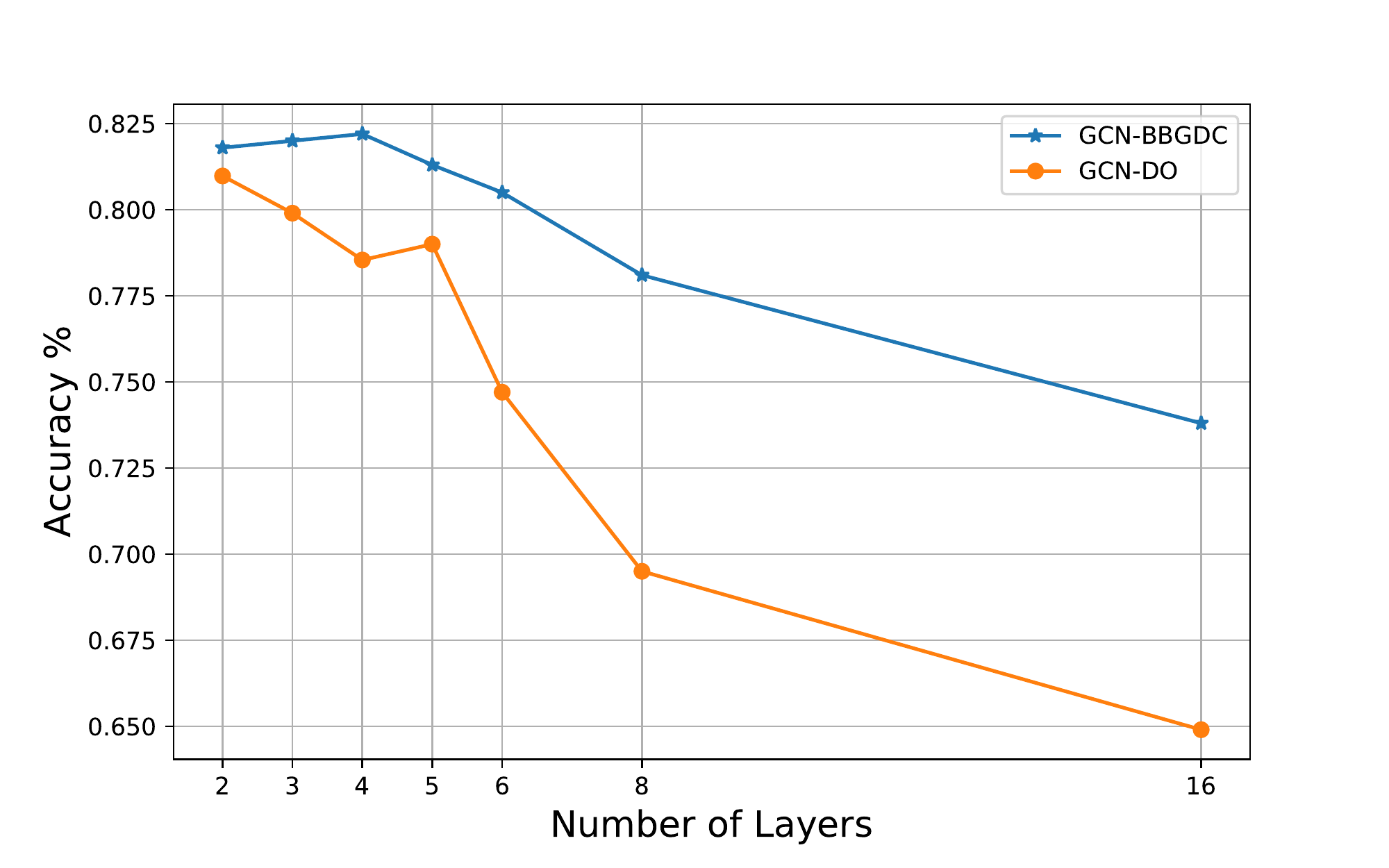}
    \end{subfigure}
    \vspace{-0.2in}
    \caption{From left to right: a) Total variation of the hidden layer outputs during training in a 4-layer GCN-BBGDC with 128-dimensional hidden layers and a 4-layer GCN-DO 128-dimensional hidden layers on Cora; b) Comparison of node classification accuracy for GCNs with a different number of hidden layers using different stochastic regularization methods. All of the hidden layers are 128 dimensional.}%
    \label{fig: totvar}
\end{figure*}

\subsection{Over-smoothing and Over-fitting}
To check how GDC helps alleviate over-smoothing in GCNs, we have tracked the total variation (TV) of the outputs of hidden layers during training. TV is a metric used in the graph signal processing literature to measure the smoothness of a signal defined over nodes of a graph \cite{chen2015signal}. More specifically, given a graph with the adjacency matrix $\mathbf{A}$ and a signal $\mathbf{x}$ defined over its nodes, TV is defined as $\mathrm{TV}(\mathbf{x}) = \Vert \mathbf{x} - (1/|\lambda_{max}|)\mathbf{A}\,\mathbf{x} \Vert_{2}^{2}$,
where $\lambda_{max}$ denotes the eigenvalue of $A$ with largest magnitude. Lower TV shows that the signal on adjacent nodes are closer to each other, indicating possible over-smoothing. 

We have compared the TV trajectories of the hidden layer outputs in a 4-layer GCN-BBGDC and a 4-layer GCN-DO normalized by their Frobenius norm, depicted in Figure \ref{fig: totvar}(a). It can be seen that, in GCN-DO, while the TV of the first layer is slightly increasing at each training epoch, the TV of the second hidden layer decreases during training. This, indeed, contributed to the poor performance of GCN-DO. On the contrary, the TVs of both first and second layers in GCN-BBGDC is increasing during training. Not only this robustness is due to the dropping connections in GDC framework, but also is related to its learnable drop rates.

With such promising results showing less over-smoothing with BBGDC, we further investigate how our proposed method works in deeper networks. We have checked the accuracy of GCN-BBGDC with a various number of 128-dimensional hidden layers ranging from 2 to 16. The results are shown in Figure \ref{fig: totvar}(b). The performance improves up to the GCN with 4 hidden layers and decreases after that. It is important to note that even though the performance drops by adding the 5-th layer, the degree to which it decreases is far less than competing methods. For example, the node classification accuracy with GCN-DO quickly drops to 69.50\% and 64.5\% with 8 and 16 layers. In addition, we should mention that the performance of GCN-DO only improves from two to three layers. This, indeed, proves GDC is a better stochastic regularization framework for GNNs in alleviating over-fitting and over-smoothing, enabling possible directions to develop deeper GNNs.

\begin{table}[!b]
   \caption{Accuracy of 128-dimensional 4-layer GCN-BBGDC with different number of blocks on Cora in semi-supervised node classification.}

  \centering
   \resizebox{0.9\columnwidth}{!}{
  \begin{tabular}{@{}l | c c c }
    \toprule
     {\textbf{Method}} & {2 blocks} & {16 blocks} & {32 blocks}\\
    \midrule
    \textbf{GCN-BBGDC} & $82.2$ & $83.0$ & $83.3$\\
    \bottomrule
  \end{tabular}
   }
  \label{tab: acc_nblock}
\end{table}
\subsection{Effect of Number of Blocks}
In GDC for every pair of input and output features, a separate mask for the adjacency matrix should be drawn. However, as we discussed in Section \ref{sec:samp_comp}, this demands large memory space. We circumvented this problem by drawing a single mask for a block of features. While we used only two blocks in our experiments presented so far, we here investigate the effect of the number of blocks on the node classification accuracy. The performance of 128-dimensional 4-layer GCN-BBGDC with 2, 16, and 32 blocks are shown in Table~\ref{tab: acc_nblock}. As can be seen, the accuracy improves as the number of blocks increases. This is due to the fact that increasing the number of blocks increases the flexibility of GDC. The choice of the number of blocks is a factor to consider for the trade off between the performance and memory usage as well as computational complexity.

\section{Conclusion}
In this paper, we proposed a unified framework for adaptive connection sampling in GNNs that generalizes existing stochastic regularization techniques for training GNNs. Our proposed method, Graph DropConnect (GDC), not only alleviates over-smoothing and over-fitting tendencies of deep GNNs, but also enables learning with uncertainty in graph analytic tasks with GNNs. %
Instead of using fixed sampling rates, 
our GDC technique parameters can be trained jointly with
GNN model parameters.
We further 
show that training a GNN with GDC is equivalent to an approximation of training Bayesian GNNs.  %
Our experimental results shows that GDC boosts the performance of GNNs in semi-supervised classification task by alleviating over-smoothing and over-fitting. We further show that the quality of uncertainty derived by GDC is better than DropOut in GNNs.

\section*{Acknowledgements}
The presented materials are based upon the work supported by the National Science Foundation under Grants ENG-1839816, IIS-1848596, CCF-1553281, IIS-1812641, IIS-1812699, and CCF-1934904.

\clearpage
\appendix

\twocolumn[
\icmltitle{Bayesian Graph Neural Networks with Adaptive Connection Sampling: Supplementary Materials}

\icmlsetsymbol{equal}{*}

\begin{icmlauthorlist}
\icmlauthor{Arman Hasanzadeh}{equal,to}
\icmlauthor{Ehsan Hajiramezanali}{equal,to}
\icmlauthor{Shahin Boluki}{to}
\icmlauthor{Mingyuan Zhou}{ed}
\icmlauthor{Nick Duffield}{to}
\icmlauthor{Krishna Narayanan}{to}
\icmlauthor{Xiaoning Qian}{to}
\end{icmlauthorlist}

\icmlaffiliation{to}{Electrical and Computer Engineering Department, Texas A\&M University, College Station, Texas, USA}
\icmlaffiliation{ed}{McCombs School of Business, The University of Texas at Austin, Austin, Texas, USA}

\icmlkeywords{Machine Learning, ICML}

\vskip 0.3in
]

\printAffiliationsAndNotice{\icmlEqualContribution}

In this supplement, we first provide an ablation study on local GDC. Dataset statistics, and further implementation details are also presented. Finally, schematics of different stochastic regularization techniques for GCNs are provided.

\section{Ablation Study: Global versus Local}
We further investigate our learnable GDC, in which for each edge at each layer a different connection sampling distribution is learned. We refer to this scenario as the \textit{local} learnable GDC. This, indeed, is a more general case than learning a single distribution for all edges in a layer. 
Expanding the variational beta-Bernoulli GDC to local learnable GDC is straightforward. Note that the KL term in the loss function can be derived in the same manner as in the global learnable GDC -- as described in Section 4 of the paper -- except that it will include the sum of $\mathrm{num\_layers} \times \mathrm{num\_edges}$ terms as opposed to the $\mathrm{num\_layers}$ terms in the global GDC.

By training the aforementioned model on the citation datasets, we find that the accuracy degrades and the KL divergence reduces to zero for every choice of prior. This phenomenon, which is known as \textit{posterior collapse} or \textit{KL vanishing}, is a common problem in variational auto-encoders for language modeling \citep{bowman2015generating, goyal2017zforcing,liu2019cyclical}. It is often due to over-parametrization in the model, which is indeed the case in the local learnable GDC. A solution to this issue could be making the parameters of the distribution dependent on the graph topology and/or node attributes. We leave this for future studies.

\section{Datasets and Implementation Details}
All of the models are implemented in PyTorch \citep{paszke2017automatic}. All of the simulations are conducted on a single NVIDIA GeForce RTX 2080 GPU node. We evaluate our proposed methods, GCN-BBDE and GCN-BBGDC, and baselines on three standard citation network benchmark datasets. We preprocess and split the dataset as done in \cite{kipf2016semi} and \cite{bojchevski2018deep}. For Cora and Cora-ML, we use 140 nodes for training, 500 nodes for validation and 1000 nodes for testing. For Citeseer, we use 120 nodes for training and the same number of nodes as Cora for validation and testing. Table~\ref{tab:stats} provides the detailed statistics of the graph datasets used in our experiments. The warm-up factor used in GCN-BBGDC with more than 2 layers for Cora and Cora-ML is $\mathrm{min}(\{1,\, \mathrm{epoch}/20\})$, and for Citeseer is $\mathrm{min}(\{1,\, \mathrm{epoch}/40\})$. We have deployed Adam optimizer \citep{kingma2014adam} in all of our experiments.

\begin{table}[!t]
    \centering
    \caption{Graph dataset statistics.}
    \vspace{0.25em}
    \resizebox{1.0\columnwidth}{!}{
    \begin{tabular}{@{}l | c c c c}
    \toprule
    Dataset & \# Classes & \# Nodes & \# Edges & \# Features\\ \midrule
    \textbf{Cora} & 7 & 2,708 & 5,429 & 1,433\\
    \textbf{Citeseer} & 3 & 3,327 & 4,732 & 3,703\\
    \textbf{Cora-ML} & 7 & 2,995 & 8,416 & 2,879\\
    \bottomrule
    \end{tabular}
    }
    \label{tab:stats}
\end{table}

\section{GDC versus Other Stochastic Regularization Techniques}
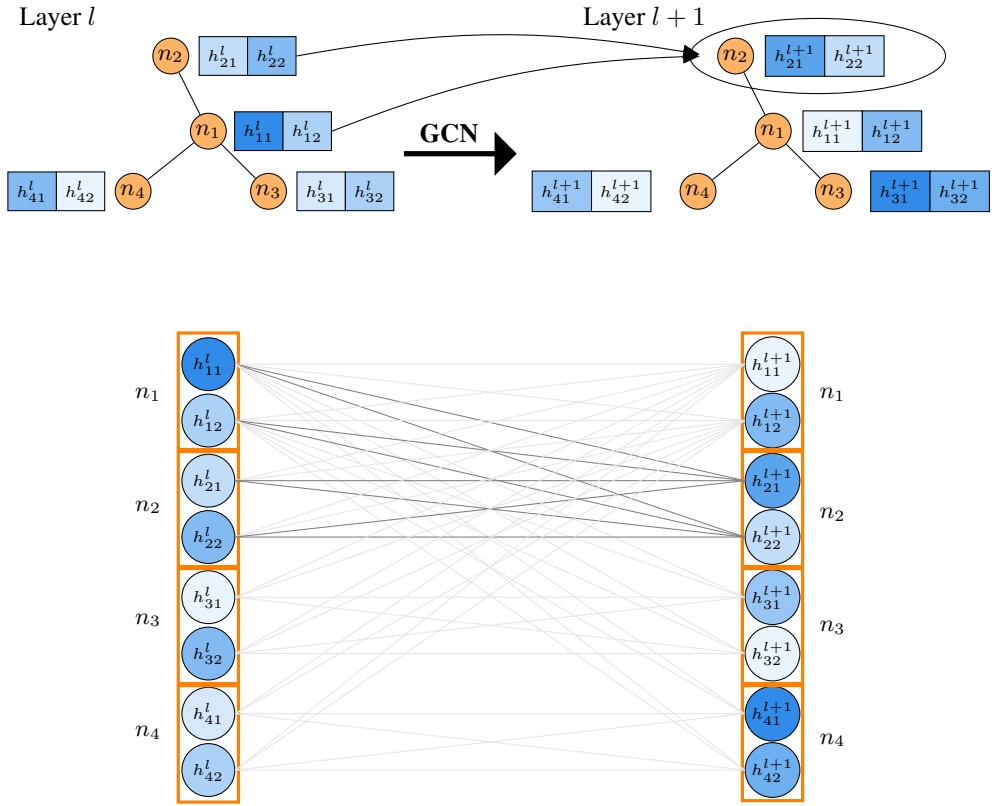
\begin{figure*}[!h]
	\centering

	\vspace{0.25cm}
	\begin{tikzpicture}[start chain=1 going right, start chain=2 going right, start chain=3 going left, start chain=4 going right, start chain=5 going right, start chain=6 going right, start chain=7 going right, start chain=8 going right, start chain=9 going right, start chain=10 going right, start chain=11 going right,start chain=12 going right,start chain=13 going right,start chain=14 going right,start chain=15 going below,node distance=-0.15mm,start chain=16 going below,start chain=17 going below,start chain=18 going below
	,start chain=19 going right
	,start chain=20 going right
	,start chain=21 going right
	,start chain=22 going right
	]

		\node[obs, minimum size=.2cm, fill=orange!60] (v1) {\small{$n_1$}};
		\node[obs, yshift=1cm, xshift=-.5cm, minimum size=.2cm, fill=orange!60] (v2) {\small{$n_2$}};
		\node[obs, yshift=-.8cm, xshift=.8cm, minimum size=.2cm, fill=orange!60] (v3) {\small{$n_3$}};
		\node[obs, yshift=-.8cm, xshift=-1cm, minimum size=.2cm, fill=orange!60] (v4) {\small{$n_4$}};
		
		\node[minimum size=.2cm, yshift=-3.5cm,xshift=-.8cm] (v1p) {\small{$n_1$}};
		\node[minimum size=.2cm,  yshift=-5.0cm,xshift=-.8cm] (v2p) {\small{$n_2$}};
		\node[minimum size=.2cm,  yshift=-6.5cm,xshift=-.8cm] (v3p) {\small{$n_3$}};
		\node[minimum size=.2cm, yshift=-8cm,xshift=-.8cm] (v4p) {\small{$n_4$}};

		\node [on chain=1,minimum size=.5cm,xshift=.1cm] (c2l1) {};
		\node [draw,on chain=1, fill=bleudefrance] (v11) {\tiny{$h_{11}^l$}};
		\node [draw,on chain=1, fill=bleudefrance!40] (v12) {\tiny{$h_{12}^l$}};
		
		\node [draw,on chain=2, fill=bleudefrance!30, yshift=1cm, xshift=.2cm] (v21) {\tiny{$h_{21}^l$}};
		\node [draw,on chain=2, fill=bleudefrance!60] (v22) {\tiny{$h_{22}^l$}};
		
		\node [draw,on chain=3, fill=bleudefrance!10, yshift=-.8cm, xshift=-1.7cm] (v41) {\tiny{$h_{42}^l$}};
		\node [draw,on chain=3, fill=bleudefrance!60] (v42) {\tiny{$h_{41}^l$}};
		
		\node [draw,on chain=4, fill=bleudefrance!20, yshift=-.8cm, xshift=1.5cm] (v31) {\tiny{$h_{31}^l$}};
		\node [draw,on chain=4, fill=bleudefrance!40] (v32) {\tiny{$h_{32}^l$}};

		\node[obs, minimum size=.3cm, fill=orange!60, left=of v1, xshift=9cm] (m1) {\small{$n_1$}};
		\node[obs, left=of v2, xshift=9cm, minimum size=.3cm, fill=orange!60] (m2) {\small{$n_2$}};
		\node[obs, left=of v3, minimum size=.3cm, fill=orange!60, xshift=9cm] (m3) {\small{$n_3$}};
		\node[obs, left=of v4, xshift=9cm, minimum size=.3cm, fill=orange!60] (m4) {\small{$n_4$}};
		\node[left=of m2] (m8) {};
		\node[left=of m1] (m9) {};
		
		\node[minimum size=.2cm,  yshift=-3.5cm,xshift=8.3cm] (m1p) {\small{$n_1$}};
		\node[minimum size=.2cm,  yshift=-5.1cm,xshift=8.3cm] (m2p) {\small{$n_2$}};
		\node[minimum size=.2cm,  yshift=-6.6cm,xshift=8.3cm] (m3p) {\small{$n_3$}};
		\node[minimum size=.2cm, yshift=-8.1cm,xshift=8.3cm] (m4p) {\small{$n_4$}};

		\node [draw,on chain=5, fill=bleudefrance!10, xshift=8.3cm] (m11) {\tiny{$h_{11}^{l+1}$}};
		\node [draw,on chain=5, fill=bleudefrance!60] (m12) {\tiny{$h_{12}^{l+1}$}};
		
		\node [draw,on chain=6, fill=bleudefrance!80, yshift=1cm, xshift=7.8cm] (m21) {\tiny{$h_{21}^{l+1}$}};
		\node [draw,on chain=6, fill=bleudefrance!30] (m22) {\tiny{$h_{22}^{l+1}$}};
		
		\node [draw,on chain=7, fill=bleudefrance!50, yshift=-.8cm, xshift=4.7cm] (m41) {\tiny{$h_{41}^{l+1}$}};
		\node [draw,on chain=7, fill=bleudefrance!10] (m42) {\tiny{$h_{42}^{l+1}$}};
		
		\node [draw,on chain=8, fill=bleudefrance, yshift=-.8cm, xshift=9.2cm] (m31) {\tiny{$h_{31}^{l+1}$}};
		\node [draw,on chain=8, fill=bleudefrance!70] (m32) {\tiny{$h_{32}^{l+1}$}};

        \path (v1) edge [nand] (v2);
        \path (v1) edge [nand] (v3);
        \path (v1) edge [nand] (v4);
        \path (m1) edge [nand] (m2);
        \path (m1) edge [nand] (m3);
        \path (m1) edge [nand] (m4);

        \draw (8.1cm,1cm) ellipse (1.7cm and .5cm);
        \path (v12.east) edge [and, bend left=10] (m8);
        \path (v22.east) edge [and, bend left=10] (m8);
		
		\node[yshift=1.5cm, xshift=-2cm] (layer1) {Layer $l$};
        \node[yshift=1.5cm, xshift=5.8cm] (layer1) {Layer $l+1$};
        
        \draw[
        -triangle 90,
        line width=.7mm,
        postaction={line width=1.2cm, shorten >=1cm, -}
    ] (2.6cm,-.3cm) -- (4.1cm,-.3cm) node[above,xshift=-.9cm] (aaa){{\bf GCN}};

        \node [obs,yshift=-3.1cm, fill=bleudefrance] (vec1) {\tiny{$h_{11}^{l}$}};
		\node [obs,yshift=-3.85cm, fill=bleudefrance!40] (vec2) {\tiny{$h_{12}^{l}$}};
		\node [obs,yshift=-4.65cm, fill=bleudefrance!30] (vec3) {\tiny{$h_{21}^{l}$}};
		\node [obs,yshift=-5.4cm, fill=bleudefrance!60] (vec4) {\tiny{$h_{22}^{l}$}};
		\node [obs,yshift=-6.2cm, fill=bleudefrance!10] (vec5) {\tiny{$h_{31}^{l}$}};
		\node [obs,yshift=-6.95cm, fill=bleudefrance!60] (vec6) {\tiny{$h_{32}^{l}$}};
		\node [obs,yshift=-7.75cm, fill=bleudefrance!20] (vec7) {\tiny{$h_{41}^{l}$}};
		\node [obs,yshift=-8.5cm, fill=bleudefrance!40] (vec8) {\tiny{$h_{42}^{l}$}};
    
	    \node [draw,on chain=18, yshift=-3.46cm,minimum width=.8cm,minimum height=1.55cm, very thick,color=orange] (mm1) {};
		\node [draw,on chain=18, minimum width=.8cm,minimum height=1.53cm, very thick,color=orange] (mm2) {};
		\node [draw,on chain=18, minimum width=.8cm,minimum height=1.53cm, very thick,color=orange] (mm3) {};
		\node [draw,on chain=18, minimum width=.8cm,minimum height=1.55cm, very thick,color=orange] (mm4) {};
		
		\node [obs, yshift=-3.1cm,xshift=7.5cm, minimum width=.6cm,minimum height=.2cm, fill=bleudefrance!10] (mec1) {\tiny{$h_{11}^{l+1}$}};
		\node [obs, yshift=-3.85cm,xshift=7.5cm,minimum width=.6cm,minimum height=.2cm, fill=bleudefrance!60] (mec2) {\tiny{$h_{12}^{l+1}$}};
		\node [obs, yshift=-4.65cm,xshift=7.5cm,minimum width=.6cm,minimum height=.2cm, fill=bleudefrance!80] (mec3) {\tiny{$h_{21}^{l+1}$}};
		\node [obs, yshift=-5.4cm,xshift=7.5cm,minimum width=.6cm,minimum height=.2cm, fill=bleudefrance!30] (mec4) {\tiny{$h_{22}^{l+1}$}};
		\node [obs, yshift=-6.2cm,xshift=7.5cm,minimum width=.6cm,minimum height=.2cm, fill=bleudefrance!50] (mec5) {\tiny{$h_{31}^{l+1}$}};
		\node [obs, yshift=-6.95cm,xshift=7.5cm,minimum width=.6cm,minimum height=.2cm, fill=bleudefrance!10] (mec6) {\tiny{$h_{32}^{l+1}$}};
		\node [obs, yshift=-7.75cm,xshift=7.5cm,minimum width=.6cm,minimum height=.2cm, fill=bleudefrance] (mec7) {\tiny{$h_{41}^{l+1}$}};
		\node [obs, yshift=-8.5cm,xshift=7.5cm,minimum width=.6cm,minimum height=.2cm, fill=bleudefrance!70] (mec8) {\tiny{$h_{42}^{l+1}$}};
		
		\node [draw,on chain=15, yshift=-3.46cm,xshift=7.5cm ,minimum width=.8cm,minimum height=1.55cm, very thick,color=orange] (mm1) {};
		\node [draw,on chain=15,, minimum width=.8cm,minimum height=1.53cm, very thick,color=orange] (mm2) {};
		\node [draw,on chain=15, minimum width=.8cm,minimum height=1.53cm, very thick,color=orange] (mm3) {};
		\node [draw,on chain=15, minimum width=.8cm,minimum height=1.53cm, very thick,color=orange] (mm4) {};
		
		\path (vec1.east) edge [nand,color=gray!20] (mec1.west);
		\path (vec1.east) edge [nand,color=gray!20] (mec2.west);
		\path (vec1.east) edge [nand,color=gray] (mec3.west);
		\path (vec1.east) edge [nand,color=gray] (mec4.west);
		\path (vec1.east) edge [nand,color=gray!20] (mec5.west);
		\path (vec1.east) edge [nand,color=gray!20] (mec6.west);
		\path (vec1.east) edge [nand,color=gray!20] (mec7.west);
		\path (vec1.east) edge [nand,color=gray!20] (mec8.west);
		
		\path (vec2.east) edge [nand,color=gray!20] (mec1.west);
		\path (vec2.east) edge [nand,color=gray!20] (mec2.west);
		\path (vec2.east) edge [nand,color=gray] (mec3.west);
		\path (vec2.east) edge [nand,color=gray] (mec4.west);
		\path (vec2.east) edge [nand,color=gray!20] (mec5.west);
		\path (vec2.east) edge [nand,color=gray!20] (mec6.west);
		\path (vec2.east) edge [nand,color=gray!20] (mec7.west);
		\path (vec2.east) edge [nand,color=gray!20] (mec8.west);
		
		\path (vec3.east) edge [nand,color=gray!20] (mec1.west);
		\path (vec3.east) edge [nand,color=gray!20] (mec2.west);
		\path (vec3.east) edge [nand,color=gray] (mec3.west);
		\path (vec3.east) edge [nand,color=gray] (mec4.west);

		\path (vec4.east) edge [nand,color=gray!20] (mec1.west);
		\path (vec4.east) edge [nand,color=gray!20] (mec2.west);
		\path (vec4.east) edge [nand,color=gray] (mec3.west);
		\path (vec4.east) edge [nand,color=gray] (mec4.west);

		\path (vec5.east) edge [nand,color=gray!20] (mec1.west);
		\path (vec5.east) edge [nand,color=gray!20] (mec2.west);
		\path (vec5.east) edge [nand,color=gray!20] (mec5.west);
		\path (vec5.east) edge [nand,color=gray!20] (mec6.west);
		
		\path (vec6.east) edge [nand,color=gray!20] (mec1.west);
		\path (vec6.east) edge [nand,color=gray!20] (mec2.west);
		\path (vec6.east) edge [nand,color=gray!20] (mec5.west);
		\path (vec6.east) edge [nand,color=gray!20] (mec6.west);

		\path (vec7.east) edge [nand,color=gray!20] (mec1.west);
		\path (vec7.east) edge [nand,color=gray!20] (mec2.west);
		\path (vec7.east) edge [nand,color=gray!20] (mec7.west);
		\path (vec7.east) edge [nand,color=gray!20] (mec8.west);

		\path (vec8.east) edge [nand,color=gray!20] (mec1.west);
		\path (vec8.east) edge [nand,color=gray!20] (mec2.west);
		\path (vec8.east) edge [nand,color=gray!20] (mec7.west);
		\path (vec8.east) edge [nand,color=gray!20] (mec8.west);

\end{tikzpicture}
	\caption{\textbf{Top}: Schematic of a GCN layer on a graph with 4 nodes. Number of both input and output features are two. The connections are localized as explicitly depicted for node 2. \textbf{Bottom}: The same GCN layer shown in a more conventional way, i.e. each layer is a vector of neurons or features. Each circle is a feature and each square represents a node. The connections are sparse and the sparsity is based on the input graph topology. The connections for node 2 in layer $l+1$ are highlighted.}
	\label{fig:gcn}
\end{figure*}

To further clarify the differences of our proposed GDC from existing stochastic regularization techniques, we draw the schematics of a GCN layer to which DropOut, DropEdge, Node Sampling, and our GDC are applied; shown in figures below. The input graph topology for the GCN layer is depicted in \ref{fig:gcn}. The number of input and output features are both two in this toy example.

\begin{figure*}[!h]
	\centering
	\vspace{0.25cm}
	\begin{tikzpicture}[start chain=1 going below,start chain=2 going below]

	    \node [obs, fill=bleudefrance] (vec1) {\tiny{$h_{11}^{l}$}};

		\node [obs,yshift=-.75cm, fill=bleudefrance!40] (vec2) {\tiny{$h_{12}^{l}$}};
		\node [obs,yshift=-1.55cm, fill=bleudefrance!30] (vec3) {\tiny{$h_{21}^{l}$}};
		\node [obs,yshift=-2.3cm, fill=bleudefrance!60] (vec4) {\tiny{$h_{22}^{l}$}};
		\node [obs,yshift=-3.1cm, fill=bleudefrance!10] (vec5) {\tiny{$h_{31}^{l}$}};
		\node [obs,yshift=-3.85cm, fill=bleudefrance!60] (vec6) {\tiny{$h_{32}^{l}$}};
		\node [obs,yshift=-4.65cm, fill=bleudefrance!20] (vec7) {\tiny{$h_{41}^{l}$}};
		\node [obs,yshift=-5.4cm, fill=bleudefrance!40] (vec8) {\tiny{$h_{42}^{l}$}};

	    \node [draw, yshift=-.36cm,minimum width=.8cm,minimum height=1.55cm, very thick,color=orange] (vv1) {};
		\node [draw,minimum width=.8cm,minimum height=1.53cm, very thick,color=orange,below=of vv1, yshift=1.02cm] (vv2) {};
		\node [draw,minimum width=.8cm,minimum height=1.53cm, very thick,color=orange,below=of vv2,yshift=1.02cm] (vv3) {};
		\node [draw,minimum width=.8cm,minimum height=1.55cm, very thick,color=orange,below=of vv3, yshift=1.02cm] (vv4) {};
		
		\node [obs, xshift=7.5cm, minimum width=.6cm,minimum height=.2cm, fill=bleudefrance!10] (mec1) {\tiny{$h_{11}^{l+1}$}};
		\node [obs, yshift=-.75cm,xshift=7.5cm,minimum width=.6cm,minimum height=.2cm, fill=bleudefrance!60] (mec2) {\tiny{$h_{12}^{l+1}$}};
		\node [obs, yshift=-1.55cm,xshift=7.5cm,minimum width=.6cm,minimum height=.2cm, fill=bleudefrance!80] (mec3) {\tiny{$h_{21}^{l+1}$}};
		\node [obs, yshift=-2.3cm,xshift=7.5cm,minimum width=.6cm,minimum height=.2cm, fill=bleudefrance!30] (mec4) {\tiny{$h_{22}^{l+1}$}};
		\node [obs, yshift=-3.1cm,xshift=7.5cm,minimum width=.6cm,minimum height=.2cm, fill=bleudefrance!50] (mec5) {\tiny{$h_{31}^{l+1}$}};
		\node [obs, yshift=-3.85cm,xshift=7.5cm,minimum width=.6cm,minimum height=.2cm, fill=bleudefrance!10] (mec6) {\tiny{$h_{32}^{l+1}$}};
		\node [obs, yshift=-4.65cm,xshift=7.5cm,minimum width=.6cm,minimum height=.2cm, fill=bleudefrance] (mec7) {\tiny{$h_{41}^{l+1}$}};
		\node [obs, yshift=-5.4cm,xshift=7.5cm,minimum width=.6cm,minimum height=.2cm, fill=bleudefrance!70] (mec8) {\tiny{$h_{42}^{l+1}$}};

		\node [draw, yshift=-.36cm,xshift=7.5cm ,minimum width=.8cm,minimum height=1.55cm, very thick,color=orange] (mm1) {};
		\node [draw, below=of mm1, yshift=1.02cm, minimum width=.8cm,minimum height=1.53cm, very thick,color=orange] (mm2) {};
		\node [draw,below=of mm2, yshift=1.02cm, minimum width=.8cm,minimum height=1.53cm, very thick,color=orange] (mm3) {};
		\node [draw,below=of mm3, yshift=1.02cm,minimum height=1.53cm, minimum width=.8cm,very thick,color=orange] (mm4) {};
		
		\node[yshift=1cm] (l1) {Layer $l$};
	    \node[yshift=1cm,xshift=7.5cm] (l2) {Layer $l+1$};
	    
		\node[minimum size=.2cm, yshift=-.45cm,xshift=-.8cm] (v1p) {\small{$n_1$}};
		\node[minimum size=.2cm,  yshift=-1.9cm,xshift=-.8cm] (v2p) {\small{$n_2$}};
		\node[minimum size=.2cm,  yshift=-3.45cm,xshift=-.8cm] (v3p) {\small{$n_3$}};
		\node[minimum size=.2cm, yshift=-4.95cm,xshift=-.8cm] (v4p) {\small{$n_4$}};
		
		\node[minimum size=.2cm, yshift=-.45cm,xshift=8.5cm] (v1p) {\small{$n_1$}};
		\node[minimum size=.2cm,  yshift=-1.9cm,xshift=8.5cm] (v2p) {\small{$n_2$}};
		\node[minimum size=.2cm,  yshift=-3.45cm,xshift=8.5cm] (v3p) {\small{$n_3$}};
		\node[minimum size=.2cm, yshift=-4.95cm,xshift=8.5cm] (v4p) {\small{$n_4$}};
		
		\path (vec1.east) edge [nand,color=gray!60] (mec1.west);
		\path (vec1.east) edge [nand,color=gray!60] (mec2.west);
		\path (vec1.east) edge [nand,color=purple!60, dashed] (mec3.west);
		\path (vec1.east) edge [nand,color=gray!60] (mec4.west);
		\path (vec1.east) edge [nand,color=purple!60,dashed] (mec5.west);
		\path (vec1.east) edge [nand,color=purple!60,dashed] (mec6.west);
		\path (vec1.east) edge [nand,color=gray!60] (mec7.west);
		\path (vec1.east) edge [nand,color=gray!60] (mec8.west);
		
		\path (vec2.east) edge [nand,color=gray!60] (mec1.west);
		\path (vec2.east) edge [nand,color=gray!60] (mec2.west);
		\path (vec2.east) edge [nand,color=gray!60] (mec3.west);
		\path (vec2.east) edge [nand,color=purple!60,dashed] (mec4.west);
		\path (vec2.east) edge [nand,color=gray!60] (mec5.west);
		\path (vec2.east) edge [nand,color=gray!60] (mec6.west);
		\path (vec2.east) edge [nand,color=gray!60] (mec7.west);
		\path (vec2.east) edge [nand,color=purple!60,dashed] (mec8.west);
		
		\path (vec3.east) edge [nand,color=purple!60, dashed] (mec1.west);
		\path (vec3.east) edge [nand,color=gray!60] (mec2.west);
		\path (vec3.east) edge [nand,color=gray!60] (mec3.west);
		\path (vec3.east) edge [nand,color=gray!60] (mec4.west);

		\path (vec4.east) edge [nand,color=gray!60] (mec1.west);
		\path (vec4.east) edge [nand,color=gray!60] (mec2.west);
		\path (vec4.east) edge [nand,color=gray!60] (mec3.west);
		\path (vec4.east) edge [nand,color=purple!60,dashed] (mec4.west);

		\path (vec5.east) edge [nand,color=gray!60] (mec1.west);
		\path (vec5.east) edge [nand,color=purple!60,dashed] (mec2.west);
		\path (vec5.east) edge [nand,color=gray!60] (mec5.west);
		\path (vec5.east) edge [nand,color=gray!60] (mec6.west);
		
		\path (vec6.east) edge [nand,color=purple!60,dashed] (mec1.west);
		\path (vec6.east) edge [nand,color=gray!60] (mec2.west);
		\path (vec6.east) edge [nand,color=gray!60] (mec5.west);
		\path (vec6.east) edge [nand,color=purple!60,dashed] (mec6.west);

		\path (vec7.east) edge [nand,color=gray!60] (mec1.west);
		\path (vec7.east) edge [nand,color=gray!60] (mec2.west);
		\path (vec7.east) edge [nand,color=gray!60] (mec7.west);
		\path (vec7.east) edge [nand,color=gray!60] (mec8.west);

		\path (vec8.east) edge [nand,color=gray!60] (mec1.west);
		\path (vec8.east) edge [nand,color=purple!60,dashed] (mec2.west);
		\path (vec8.east) edge [nand,color=purple!60,dashed] (mec7.west);
		\path (vec8.east) edge [nand,color=gray!60] (mec8.west);

\end{tikzpicture}
\caption{Schematic of our proposed GDC. Each circle is a feature and each square represents a node. GDC drops connections independently across layers. The dashed lines show dropped connections and the gray ones show the kept connections.}
	\label{fig:gdc}
\end{figure*}
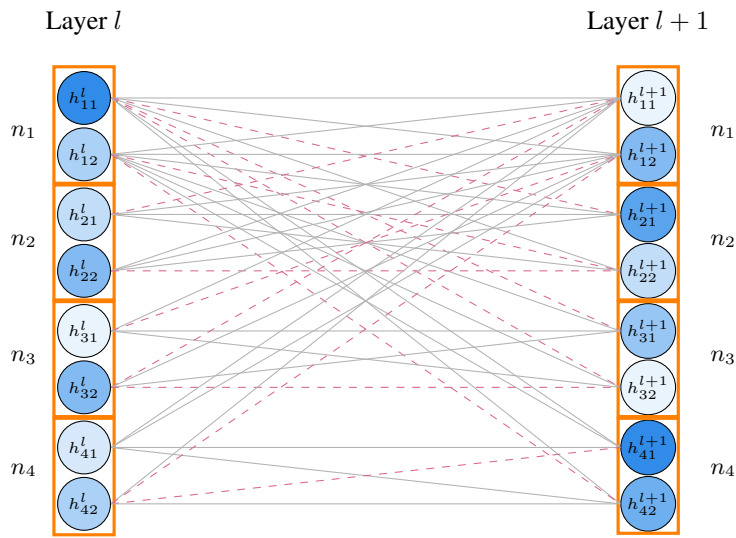

\clearpage

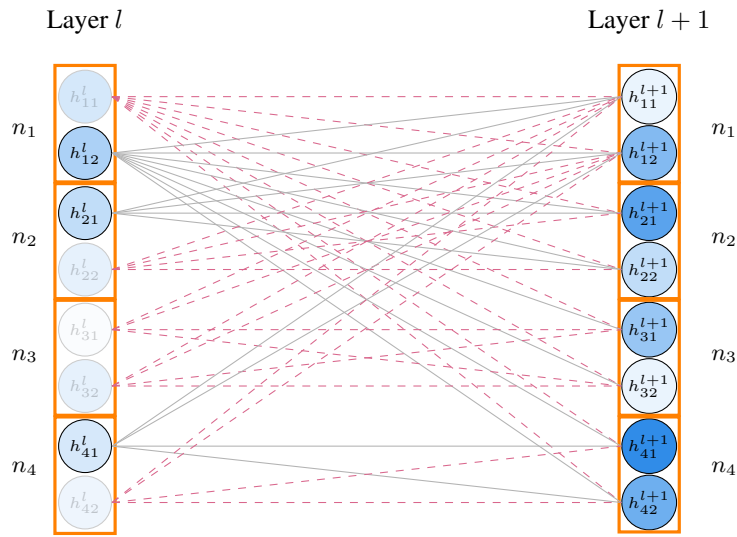
\begin{figure*}[!h]
	\centering
	\vspace{0.25cm}
	\begin{tikzpicture}[start chain=1 going below,start chain=2 going below]

	    \node [obs, fill=bleudefrance,opacity=.2] (vec1) {\tiny{$h_{11}^{l}$}};
		\node [obs,yshift=-.75cm, fill=bleudefrance!40] (vec2) {\tiny{$h_{12}^{l}$}};
		\node [obs,yshift=-1.55cm, fill=bleudefrance!30] (vec3) {\tiny{$h_{21}^{l}$}};
		\node [obs,yshift=-2.3cm, fill=bleudefrance!60,opacity=.2] (vec4) {\tiny{$h_{22}^{l}$}};
		\node [obs,yshift=-3.1cm, fill=bleudefrance!10,opacity=.2] (vec5) {\tiny{$h_{31}^{l}$}};
		\node [obs,yshift=-3.85cm, fill=bleudefrance!60,opacity=.2] (vec6) {\tiny{$h_{32}^{l}$}};
		\node [obs,yshift=-4.65cm, fill=bleudefrance!20] (vec7) {\tiny{$h_{41}^{l}$}};
		\node [obs,yshift=-5.4cm, fill=bleudefrance!40,opacity=.2] (vec8) {\tiny{$h_{42}^{l}$}};

        \node[minimum size=.2cm, yshift=-.45cm,xshift=-.8cm] (v1p) {\small{$n_1$}};
		\node[minimum size=.2cm,  yshift=-1.9cm,xshift=-.8cm] (v2p) {\small{$n_2$}};
		\node[minimum size=.2cm,  yshift=-3.45cm,xshift=-.8cm] (v3p) {\small{$n_3$}};
		\node[minimum size=.2cm, yshift=-4.95cm,xshift=-.8cm] (v4p) {\small{$n_4$}};
    
	    \node [draw, yshift=-.36cm,minimum width=.8cm,minimum height=1.55cm, very thick,color=orange] (vv1) {};
		\node [draw,minimum width=.8cm,minimum height=1.53cm, very thick,color=orange,below=of vv1, yshift=1.02cm] (vv2) {};
		\node [draw,minimum width=.8cm,minimum height=1.53cm, very thick,color=orange,below=of vv2,yshift=1.02cm] (vv3) {};
		\node [draw,minimum width=.8cm,minimum height=1.55cm, very thick,color=orange,below=of vv3, yshift=1.02cm] (vv4) {};
		
		\node [obs, xshift=7.5cm, minimum width=.6cm,minimum height=.2cm, fill=bleudefrance!10] (mec1) {\tiny{$h_{11}^{l+1}$}};
		\node [obs, yshift=-.75cm,xshift=7.5cm,minimum width=.6cm,minimum height=.2cm, fill=bleudefrance!60] (mec2) {\tiny{$h_{12}^{l+1}$}};
		\node [obs, yshift=-1.55cm,xshift=7.5cm,minimum width=.6cm,minimum height=.2cm, fill=bleudefrance!80] (mec3) {\tiny{$h_{21}^{l+1}$}};
		\node [obs, yshift=-2.3cm,xshift=7.5cm,minimum width=.6cm,minimum height=.2cm, fill=bleudefrance!30] (mec4) {\tiny{$h_{22}^{l+1}$}};
		\node [obs, yshift=-3.1cm,xshift=7.5cm,minimum width=.6cm,minimum height=.2cm, fill=bleudefrance!50] (mec5) {\tiny{$h_{31}^{l+1}$}};
		\node [obs, yshift=-3.85cm,xshift=7.5cm,minimum width=.6cm,minimum height=.2cm, fill=bleudefrance!10] (mec6) {\tiny{$h_{32}^{l+1}$}};
		\node [obs, yshift=-4.65cm,xshift=7.5cm,minimum width=.6cm,minimum height=.2cm, fill=bleudefrance] (mec7) {\tiny{$h_{41}^{l+1}$}};
		\node [obs, yshift=-5.4cm,xshift=7.5cm,minimum width=.6cm,minimum height=.2cm, fill=bleudefrance!70] (mec8) {\tiny{$h_{42}^{l+1}$}};

		\node[minimum size=.2cm, yshift=-.45cm,xshift=8.5cm] (v1p) {\small{$n_1$}};
		\node[minimum size=.2cm,  yshift=-1.9cm,xshift=8.5cm] (v2p) {\small{$n_2$}};
		\node[minimum size=.2cm,  yshift=-3.45cm,xshift=8.5cm] (v3p) {\small{$n_3$}};
		\node[minimum size=.2cm, yshift=-4.95cm,xshift=8.5cm] (v4p) {\small{$n_4$}};
		
		\node [draw, yshift=-.36cm,xshift=7.5cm ,minimum width=.8cm,minimum height=1.55cm, very thick,color=orange] (mm1) {};
		\node [draw, below=of mm1, yshift=1.02cm, minimum width=.8cm,minimum height=1.53cm, very thick,color=orange] (mm2) {};
		\node [draw,below=of mm2, yshift=1.02cm, minimum width=.8cm,minimum height=1.53cm, very thick,color=orange] (mm3) {};
		\node [draw,below=of mm3, yshift=1.02cm,minimum height=1.53cm, minimum width=.8cm,very thick,color=orange] (mm4) {};

		\node[yshift=1cm] (l1) {Layer $l$};
	    \node[yshift=1cm,xshift=7.5cm] (l2) {Layer $l+1$};
	    
		\path (vec1.east) edge [nand,color=purple!60, dashed] (mec1.west);
		\path (vec1.east) edge [nand,color=purple!60, dashed] (mec2.west);
		\path (vec1.east) edge [nand,color=purple!60, dashed] (mec3.west);
		\path (vec1.east) edge [nand,color=purple!60, dashed] (mec4.west);
		\path (vec1.east) edge [nand,color=purple!60, dashed] (mec5.west);
		\path (vec1.east) edge [nand,color=purple!60, dashed] (mec6.west);
		\path (vec1.east) edge [nand,color=purple!60, dashed] (mec7.west);
		\path (vec1.east) edge [nand,color=purple!60, dashed] (mec8.west);
		
		\path (vec2.east) edge [nand,color=gray!60] (mec1.west);
		\path (vec2.east) edge [nand,color=gray!60] (mec2.west);
		\path (vec2.east) edge [nand,color=gray!60] (mec3.west);
		\path (vec2.east) edge [nand,color=gray!60] (mec4.west);
		\path (vec2.east) edge [nand,color=gray!60] (mec5.west);
		\path (vec2.east) edge [nand,color=gray!60] (mec6.west);
		\path (vec2.east) edge [nand,color=gray!60] (mec7.west);
		\path (vec2.east) edge [nand,color=gray!60] (mec8.west);
		
		\path (vec3.east) edge [nand,color=gray!60] (mec1.west);
		\path (vec3.east) edge [nand,color=gray!60] (mec2.west);
		\path (vec3.east) edge [nand,color=gray!60] (mec3.west);
		\path (vec3.east) edge [nand,color=gray!60] (mec4.west);

		\path (vec4.east) edge [nand,color=purple!60, dashed] (mec1.west);
		\path (vec4.east) edge [nand,color=purple!60, dashed] (mec2.west);
		\path (vec4.east) edge [nand,color=purple!60, dashed] (mec3.west);
		\path (vec4.east) edge [nand,color=purple!60, dashed] (mec4.west);

		\path (vec5.east) edge [nand,color=purple!60, dashed] (mec1.west);
		\path (vec5.east) edge [nand,color=purple!60, dashed] (mec2.west);
		\path (vec5.east) edge [nand,color=purple!60, dashed] (mec5.west);
		\path (vec5.east) edge [nand,color=purple!60, dashed] (mec6.west);
		
		\path (vec6.east) edge [nand,color=purple!60, dashed] (mec1.west);
		\path (vec6.east) edge [nand,color=purple!60, dashed] (mec2.west);
		\path (vec6.east) edge [nand,color=purple!60, dashed] (mec5.west);
		\path (vec6.east) edge [nand,color=purple!60, dashed] (mec6.west);

		\path (vec7.east) edge [nand,color=gray!60] (mec1.west);
		\path (vec7.east) edge [nand,color=gray!60] (mec2.west);
		\path (vec7.east) edge [nand,color=gray!60] (mec7.west);
		\path (vec7.east) edge [nand,color=gray!60] (mec8.west);

		\path (vec8.east) edge [nand,color=purple!60, dashed] (mec1.west);
		\path (vec8.east) edge [nand,color=purple!60, dashed] (mec2.west);
		\path (vec8.east) edge [nand,color=purple!60, dashed] (mec7.west);
		\path (vec8.east) edge [nand,color=purple!60, dashed] (mec8.west);

\end{tikzpicture}
	\caption{Schematic of DropOut \citep{srivastava2014dropout}. Each circle is a feature and each square represents a node. DropOut drops features at each layer. The faded circles represent dropped features while the other ones are kept. The dashed lines show dropped connections and the gray ones show the kept ones.}
	\label{fig:dropout}
\end{figure*}

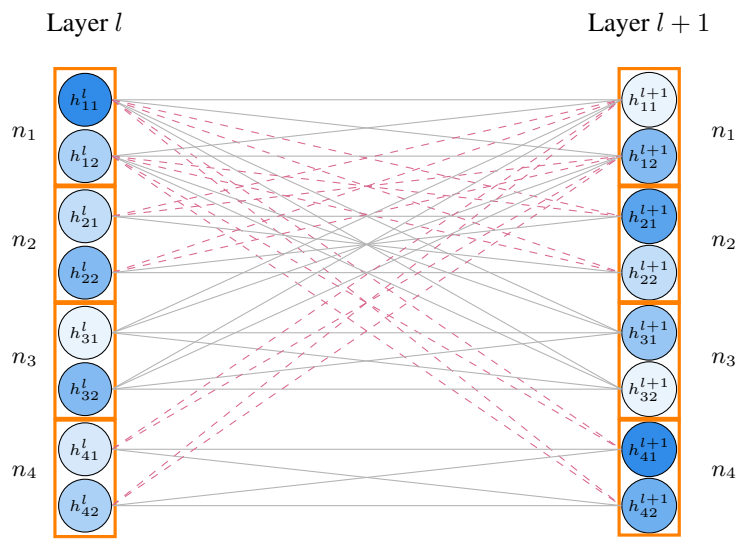
\begin{figure*}[!h]
	\centering
		\vspace{0.25cm}
\begin{tikzpicture}[start chain=1 going below,start chain=2 going below]

	    \node [obs, fill=bleudefrance] (vec1) {\tiny{$h_{11}^{l}$}};
		\node [obs,yshift=-.75cm, fill=bleudefrance!40] (vec2) {\tiny{$h_{12}^{l}$}};
		\node [obs,yshift=-1.55cm, fill=bleudefrance!30] (vec3) {\tiny{$h_{21}^{l}$}};
		\node [obs,yshift=-2.3cm, fill=bleudefrance!60] (vec4) {\tiny{$h_{22}^{l}$}};
		\node [obs,yshift=-3.1cm, fill=bleudefrance!10] (vec5) {\tiny{$h_{31}^{l}$}};
		\node [obs,yshift=-3.85cm, fill=bleudefrance!60] (vec6) {\tiny{$h_{32}^{l}$}};
		\node [obs,yshift=-4.65cm, fill=bleudefrance!20] (vec7) {\tiny{$h_{41}^{l}$}};
		\node [obs,yshift=-5.4cm, fill=bleudefrance!40] (vec8) {\tiny{$h_{42}^{l}$}};

	    \node [draw, yshift=-.36cm,minimum width=.8cm,minimum height=1.55cm, very thick,color=orange] (vv1) {};
		\node [draw,minimum width=.8cm,minimum height=1.53cm, very thick,color=orange,below=of vv1, yshift=1.02cm] (vv2) {};
		\node [draw,minimum width=.8cm,minimum height=1.53cm, very thick,color=orange,below=of vv2,yshift=1.02cm] (vv3) {};
		\node [draw,minimum width=.8cm,minimum height=1.55cm, very thick,color=orange,below=of vv3, yshift=1.02cm] (vv4) {};
		
		\node [obs, xshift=7.5cm, minimum width=.6cm,minimum height=.2cm, fill=bleudefrance!10] (mec1) {\tiny{$h_{11}^{l+1}$}};
		\node [obs, yshift=-.75cm,xshift=7.5cm,minimum width=.6cm,minimum height=.2cm, fill=bleudefrance!60] (mec2) {\tiny{$h_{12}^{l+1}$}};
		\node [obs, yshift=-1.55cm,xshift=7.5cm,minimum width=.6cm,minimum height=.2cm, fill=bleudefrance!80] (mec3) {\tiny{$h_{21}^{l+1}$}};
		\node [obs, yshift=-2.3cm,xshift=7.5cm,minimum width=.6cm,minimum height=.2cm, fill=bleudefrance!30] (mec4) {\tiny{$h_{22}^{l+1}$}};
		\node [obs, yshift=-3.1cm,xshift=7.5cm,minimum width=.6cm,minimum height=.2cm, fill=bleudefrance!50] (mec5) {\tiny{$h_{31}^{l+1}$}};
		\node [obs, yshift=-3.85cm,xshift=7.5cm,minimum width=.6cm,minimum height=.2cm, fill=bleudefrance!10] (mec6) {\tiny{$h_{32}^{l+1}$}};
		\node [obs, yshift=-4.65cm,xshift=7.5cm,minimum width=.6cm,minimum height=.2cm, fill=bleudefrance] (mec7) {\tiny{$h_{41}^{l+1}$}};
		\node [obs, yshift=-5.4cm,xshift=7.5cm,minimum width=.6cm,minimum height=.2cm, fill=bleudefrance!70] (mec8) {\tiny{$h_{42}^{l+1}$}};
		
		\node [draw, yshift=-.36cm,xshift=7.5cm ,minimum width=.8cm,minimum height=1.55cm, very thick,color=orange] (mm1) {};
		\node [draw, below=of mm1, yshift=1.02cm, minimum width=.8cm,minimum height=1.53cm, very thick,color=orange] (mm2) {};
		\node [draw,below=of mm2, yshift=1.02cm, minimum width=.8cm,minimum height=1.53cm, very thick,color=orange] (mm3) {};
		\node [draw,below=of mm3, yshift=1.02cm,minimum height=1.53cm, minimum width=.8cm,very thick,color=orange] (mm4) {};
		
		\node[yshift=1cm] (l1) {Layer $l$};
	    \node[yshift=1cm,xshift=7.5cm] (l2) {Layer $l+1$};
	    
	    \node[minimum size=.2cm, yshift=-.45cm,xshift=-.8cm] (v1p) {\small{$n_1$}};
		\node[minimum size=.2cm,  yshift=-1.9cm,xshift=-.8cm] (v2p) {\small{$n_2$}};
		\node[minimum size=.2cm,  yshift=-3.45cm,xshift=-.8cm] (v3p) {\small{$n_3$}};
		\node[minimum size=.2cm, yshift=-4.95cm,xshift=-.8cm] (v4p) {\small{$n_4$}};
	    
	    \node[minimum size=.2cm, yshift=-.45cm,xshift=8.5cm] (v1p) {\small{$n_1$}};
		\node[minimum size=.2cm,  yshift=-1.9cm,xshift=8.5cm] (v2p) {\small{$n_2$}};
		\node[minimum size=.2cm,  yshift=-3.45cm,xshift=8.5cm] (v3p) {\small{$n_3$}};
		\node[minimum size=.2cm, yshift=-4.95cm,xshift=8.5cm] (v4p) {\small{$n_4$}};
		
		\path (vec1.east) edge [nand,color=gray!60] (mec1.west);
		\path (vec1.east) edge [nand,color=gray!60] (mec2.west);
		\path (vec1.east) edge [nand,color=purple!60, dashed] (mec3.west);
		\path (vec1.east) edge [nand,color=purple!60, dashed] (mec4.west);
		\path (vec1.east) edge [nand,color=gray!60] (mec5.west);
		\path (vec1.east) edge [nand,color=gray!60] (mec6.west);
		\path (vec1.east) edge [nand,color=purple!60,dashed] (mec7.west);
		\path (vec1.east) edge [nand,color=purple!60,dashed] (mec8.west);
		
		\path (vec2.east) edge [nand,color=gray!60] (mec1.west);
		\path (vec2.east) edge [nand,color=gray!60] (mec2.west);
		\path (vec2.east) edge [nand,color=purple!60,dashed] (mec3.west);
		\path (vec2.east) edge [nand,color=purple!60,dashed] (mec4.west);
		\path (vec2.east) edge [nand,color=gray!60] (mec5.west);
		\path (vec2.east) edge [nand,color=gray!60] (mec6.west);
		\path (vec2.east) edge [nand,color=purple!60,dashed] (mec7.west);
		\path (vec2.east) edge [nand,color=purple!60,dashed] (mec8.west);
		
		\path (vec3.east) edge [nand,color=purple!60, dashed] (mec1.west);
		\path (vec3.east) edge [nand,color=purple!60, dashed] (mec2.west);
		\path (vec3.east) edge [nand,color=gray!60] (mec3.west);
		\path (vec3.east) edge [nand,color=gray!60] (mec4.west);

		\path (vec4.east) edge [nand,color=purple!60, dashed] (mec1.west);
		\path (vec4.east) edge [nand,color=purple!60, dashed] (mec2.west);
		\path (vec4.east) edge [nand,color=gray!60] (mec3.west);
		\path (vec4.east) edge [nand,color=gray!60] (mec4.west);

		\path (vec5.east) edge [nand,color=gray!60] (mec1.west);
		\path (vec5.east) edge [nand,color=gray!60] (mec2.west);
		\path (vec5.east) edge [nand,color=gray!60] (mec5.west);
		\path (vec5.east) edge [nand,color=gray!60] (mec6.west);
		
		\path (vec6.east) edge [nand,color=gray!60] (mec1.west);
		\path (vec6.east) edge [nand,color=gray!60] (mec2.west);
		\path (vec6.east) edge [nand,color=gray!60] (mec5.west);
		\path (vec6.east) edge [nand,color=gray!60] (mec6.west);

		\path (vec7.east) edge [nand,color=purple!60, dashed] (mec1.west);
		\path (vec7.east) edge [nand,color=purple!60, dashed] (mec2.west);
		\path (vec7.east) edge [nand,color=gray!60] (mec7.west);
		\path (vec7.east) edge [nand,color=gray!60] (mec8.west);

		\path (vec8.east) edge [nand,color=purple!60,dashed] (mec1.west);
		\path (vec8.east) edge [nand,color=purple!60,dashed] (mec2.west);
		\path (vec8.east) edge [nand,color=gray!60] (mec7.west);
		\path (vec8.east) edge [nand,color=gray!60] (mec8.west);

\end{tikzpicture}
\caption{Schematic of DropEdge \citep{rong2019dropedge}. Each circle is a feature and each square represents a node. DropEdge drops edges between nodes hence all of the connections between their corresponding channels are dropped. Note that the mask in DropEdge is symmetric. In this example, the edge between nodes 1 and 2 as well as the edge between nodes 1 and 4 are dropped. The dashed lines show dropped connections and the gray ones show the kept ones.}

	\label{fig:dropedge}
\end{figure*}

\clearpage

\begin{figure*}[!t]
	\centering
	\vspace{0.25cm}
		\begin{tikzpicture}[start chain=1 going below,start chain=2 going below]

	    \node [obs, fill=bleudefrance] (vec1) {\tiny{$h_{11}^{l}$}};
		\node [obs,yshift=-.75cm, fill=bleudefrance!40] (vec2) {\tiny{$h_{12}^{l}$}};
		\node [obs,yshift=-1.55cm, fill=bleudefrance!30, opacity=.2] (vec3) {\tiny{$h_{21}^{l}$}};
		\node [obs,yshift=-2.3cm, fill=bleudefrance!60, opacity=.2] (vec4) {\tiny{$h_{22}^{l}$}};
		\node [obs,yshift=-3.1cm, fill=bleudefrance!10] (vec5) {\tiny{$h_{31}^{l}$}};
		\node [obs,yshift=-3.85cm, fill=bleudefrance!60] (vec6) {\tiny{$h_{32}^{l}$}};
		\node [obs,yshift=-4.65cm, fill=bleudefrance!20, opacity=.2] (vec7) {\tiny{$h_{41}^{l}$}};
		\node [obs,yshift=-5.4cm, fill=bleudefrance!40, opacity=.2] (vec8) {\tiny{$h_{42}^{l}$}};

	    \node [draw, yshift=-.36cm,minimum width=.8cm,minimum height=1.55cm, very thick,color=orange] (vv1) {};
		\node [draw,minimum width=.8cm,minimum height=1.53cm, very thick,color=orange,below=of vv1, yshift=1.02cm,opacity=.2] (vv2) {};
		\node [draw,minimum width=.8cm,minimum height=1.53cm, very thick,color=orange,below=of vv2,yshift=1.02cm] (vv3) {};
		\node [draw,minimum width=.8cm,minimum height=1.55cm, very thick,color=orange,below=of vv3, yshift=1.02cm,,opacity=.2] (vv4) {};
		
		\node [obs, xshift=7.5cm, minimum width=.6cm,minimum height=.2cm, fill=bleudefrance!10] (mec1) {\tiny{$h_{11}^{l+1}$}};
		\node [obs, yshift=-.75cm,xshift=7.5cm,minimum width=.6cm,minimum height=.2cm, fill=bleudefrance!60] (mec2) {\tiny{$h_{12}^{l+1}$}};
		\node [obs, yshift=-1.55cm,xshift=7.5cm,minimum width=.6cm,minimum height=.2cm, fill=bleudefrance!80] (mec3) {\tiny{$h_{21}^{l+1}$}};
		\node [obs, yshift=-2.3cm,xshift=7.5cm,minimum width=.6cm,minimum height=.2cm, fill=bleudefrance!30] (mec4) {\tiny{$h_{22}^{l+1}$}};
		\node [obs, yshift=-3.1cm,xshift=7.5cm,minimum width=.6cm,minimum height=.2cm, fill=bleudefrance!50] (mec5) {\tiny{$h_{31}^{l+1}$}};
		\node [obs, yshift=-3.85cm,xshift=7.5cm,minimum width=.6cm,minimum height=.2cm, fill=bleudefrance!10] (mec6) {\tiny{$h_{32}^{l+1}$}};
		\node [obs, yshift=-4.65cm,xshift=7.5cm,minimum width=.6cm,minimum height=.2cm, fill=bleudefrance] (mec7) {\tiny{$h_{41}^{l+1}$}};
		\node [obs, yshift=-5.4cm,xshift=7.5cm,minimum width=.6cm,minimum height=.2cm, fill=bleudefrance!70] (mec8) {\tiny{$h_{42}^{l+1}$}};
		
		\node [draw, yshift=-.36cm,xshift=7.5cm ,minimum width=.8cm,minimum height=1.55cm, very thick,color=orange] (mm1) {};
		\node [draw, below=of mm1, yshift=1.02cm, minimum width=.8cm,minimum height=1.53cm, very thick,color=orange] (mm2) {};
		\node [draw,below=of mm2, yshift=1.02cm, minimum width=.8cm,minimum height=1.53cm, very thick,color=orange] (mm3) {};
		\node [draw,below=of mm3, yshift=1.02cm,minimum height=1.53cm, minimum width=.8cm,very thick,color=orange] (mm4) {};

		\node[yshift=1cm] (l1) {Layer $l$};
	    \node[yshift=1cm,xshift=7.5cm] (l2) {Layer $l+1$};
	    
	    \node[minimum size=.2cm, yshift=-.45cm,xshift=-.8cm] (v1p) {\small{$n_1$}};
		\node[minimum size=.2cm,  yshift=-1.9cm,xshift=-.8cm] (v2p) {\small{$n_2$}};
		\node[minimum size=.2cm,  yshift=-3.45cm,xshift=-.8cm] (v3p) {\small{$n_3$}};
		\node[minimum size=.2cm, yshift=-4.95cm,xshift=-.8cm] (v4p) {\small{$n_4$}};
	    
	    \node[minimum size=.2cm, yshift=-.45cm,xshift=8.5cm] (v1p) {\small{$n_1$}};
		\node[minimum size=.2cm,  yshift=-1.9cm,xshift=8.5cm] (v2p) {\small{$n_2$}};
		\node[minimum size=.2cm,  yshift=-3.45cm,xshift=8.5cm] (v3p) {\small{$n_3$}};
		\node[minimum size=.2cm, yshift=-4.95cm,xshift=8.5cm] (v4p) {\small{$n_4$}};
	    
		\path (vec1.east) edge [nand,color=gray!60] (mec1.west);
		\path (vec1.east) edge [nand,color=gray!60] (mec2.west);
		\path (vec1.east) edge [nand,color=gray!60] (mec3.west);
		\path (vec1.east) edge [nand,color=gray!60] (mec4.west);
		\path (vec1.east) edge [nand,color=gray!60] (mec5.west);
		\path (vec1.east) edge [nand,color=gray!60] (mec6.west);
		\path (vec1.east) edge [nand,color=gray!60] (mec7.west);
		\path (vec1.east) edge [nand,color=gray!60] (mec8.west);
		
		\path (vec2.east) edge [nand,color=gray!60] (mec1.west);
		\path (vec2.east) edge [nand,color=gray!60] (mec2.west);
		\path (vec2.east) edge [nand,color=gray!60] (mec3.west);
		\path (vec2.east) edge [nand,color=gray!60] (mec4.west);
		\path (vec2.east) edge [nand,color=gray!60] (mec5.west);
		\path (vec2.east) edge [nand,color=gray!60] (mec6.west);
		\path (vec2.east) edge [nand,color=gray!60] (mec7.west);
		\path (vec2.east) edge [nand,color=gray!60] (mec8.west);
		
		\path (vec3.east) edge [nand,color=purple!60, dashed] (mec1.west);
		\path (vec3.east) edge [nand,color=purple!60, dashed] (mec2.west);
		\path (vec3.east) edge [nand,color=purple!60, dashed] (mec3.west);
		\path (vec3.east) edge [nand,color=purple!60, dashed] (mec4.west);

		\path (vec4.east) edge [nand,color=purple!60, dashed] (mec1.west);
		\path (vec4.east) edge [nand,color=purple!60, dashed] (mec2.west);
		\path (vec4.east) edge [nand,color=purple!60, dashed] (mec3.west);
		\path (vec4.east) edge [nand,color=purple!60, dashed] (mec4.west);

		\path (vec5.east) edge [nand,color=gray!60] (mec1.west);
		\path (vec5.east) edge [nand,color=gray!60] (mec2.west);
		\path (vec5.east) edge [nand,color=gray!60] (mec5.west);
		\path (vec5.east) edge [nand,color=gray!60] (mec6.west);
		
		\path (vec6.east) edge [nand,color=gray!60] (mec1.west);
		\path (vec6.east) edge [nand,color=gray!60] (mec2.west);
		\path (vec6.east) edge [nand,color=gray!60] (mec5.west);
		\path (vec6.east) edge [nand,color=gray!60] (mec6.west);

		\path (vec7.east) edge [nand,color=purple!60, dashed] (mec1.west);
		\path (vec7.east) edge [nand,color=purple!60, dashed] (mec2.west);
		\path (vec7.east) edge [nand,color=purple!60, dashed] (mec7.west);
		\path (vec7.east) edge [nand,color=purple!60, dashed] (mec8.west);

		\path (vec8.east) edge [nand,color=purple!60,dashed] (mec1.west);
		\path (vec8.east) edge [nand,color=purple!60,dashed] (mec2.west);
		\path (vec8.east) edge [nand,color=purple!60, dashed] (mec7.west);
		\path (vec8.east) edge [nand,color=purple!60, dashed] (mec8.west);

\end{tikzpicture}
\caption{Schematic of the node sampling strategy in FastGCN \citep{chen2018fastgcn}. Each circle is a feature and each square represents a node. FastGCN drops nodes hence all of the connections to that node are dropped. The faded nodes represents the dropped nodes. The dashed lines show dropped connections and the gray ones show the kept ones.}
	
	\label{fig:ndsamp}
\end{figure*}
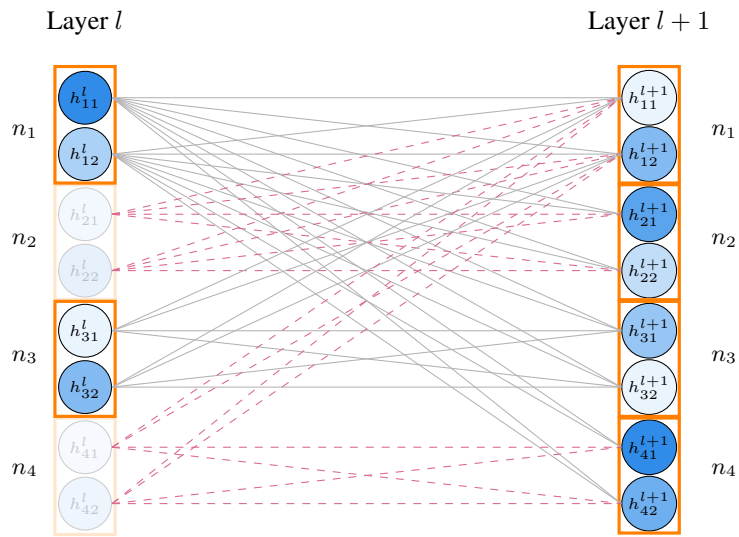

\clearpage
\bibliography{ref.bib}
\bibliographystyle{icml2020}

\end{document}